%% file: main.tex
\documentclass[conference]{IEEEtran}
\usepackage{times}

\usepackage[numbers]{natbib}
\usepackage{multicol}
\usepackage[bookmarks=true]{hyperref}
\usepackage{xcolor}
\usepackage{xspace}
\usepackage{graphicx}
\usepackage{amsmath}
\usepackage{amssymb}

\usepackage{algorithm}
\usepackage{algpseudocode}
\usepackage{booktabs}  
\usepackage{multirow}  
\usepackage{graphicx}  
\usepackage{array}
\usepackage{capt-of}
\usepackage{booktabs}
\usepackage{multirow}
\usepackage{graphicx}
\usepackage[table]{xcolor}
\usepackage{colortbl}     
\usepackage{pgfplots}
\usepackage{enumitem}
\pgfplotsset{compat=1.18}
\usepgfplotslibrary{groupplots} 
\usetikzlibrary{patterns}
\usepackage{cuted}
\usepackage{tabularx}
\definecolor{mygray}{gray}{0.9}

\newcommand{\nm}{\textbf{LiLo-VLA}\xspace}

\begin{document}

\title{LiLo-VLA: Compositional Long-Horizon Manipulation via Linked Object-Centric Policies}

% You will get a Paper-ID when submitting a pdf file to the conference system
% \author{Author Names Omitted for Anonymous Review. Paper-ID 414}

\author{
    Yue Yang$^{1, \dagger}$, Shuo Cheng$^{2}$, Yu Fang$^{1}$, Homanga Bharadhwaj$^{3}$, \\
    Mingyu Ding$^{1}$, Gedas Bertasius$^{1}$, Daniel Szafir$^{1}$ \\[1.2ex]
    \textbf{$^{1}$University of North Carolina at Chapel Hill} \quad
    \textbf{$^{2}$Georgia Institute of Technology} \quad
    \textbf{$^{3}$Carnegie Mellon University} \\[1ex]
    $^{\dagger}$Corresponding author: {\tt\small yygx@cs.unc.edu}
}

% \author{\authorblockN{Michael Shell}
% \authorblockA{School of Electrical and\\Computer Engineering\\
% Georgia Institute of Technology\\
% Atlanta, Georgia 30332--0250\\
% Email: mshell@ece.gatech.edu}
% \and
% \authorblockN{Homer Simpson}
% \authorblockA{Twentieth Century Fox\\
% Email: homer@thesimpsons.com}
% \and
% \authorblockN{James Kirk\\ and Montgomery Scott}
% \authorblockA{Starfleet Academy\\
% San Francisco, California 96678-2391\\
% Telephone: (800) 555--1212\\

% avoiding spaces at the end of the author lines is not a problem with
% conference papers because we don't use \thanks or \IEEEmembership

% \author{\authorblockN{Michael Shell\authorrefmark{1},
% Homer Simpson\authorrefmark{2},
% James Kirk\authorrefmark{3}, 
% Montgomery Scott\authorrefmark{3} and
% Eldon Tyrell\authorrefmark{4}}
% \authorblockA{\authorrefmark{1}School of Electrical and Computer Engineering\\
% Georgia Institute of Technology,
% Atlanta, Georgia 30332--0250\\ Email: mshell@ece.gatech.edu}
% \authorblockA{\authorrefmark{2}Twentieth Century Fox, Springfield, USA\\
% Email: homer@thesimpsons.com}
% \authorblockA{\authorrefmark{3}Starfleet Academy, San Francisco, California 96678-2391\\
% Telephone: (800) 555--1212, Fax: (888) 555--1212}
% \authorblockA{\authorrefmark{4}Tyrell Inc., 123 Replicant Street, Los Angeles, California 90210--4321}}

% \maketitle

\twocolumn[{%
\renewcommand\twocolumn[1][]{#1}%
\maketitle
\begin{center}
    \centering
    \includegraphics[width=\textwidth]{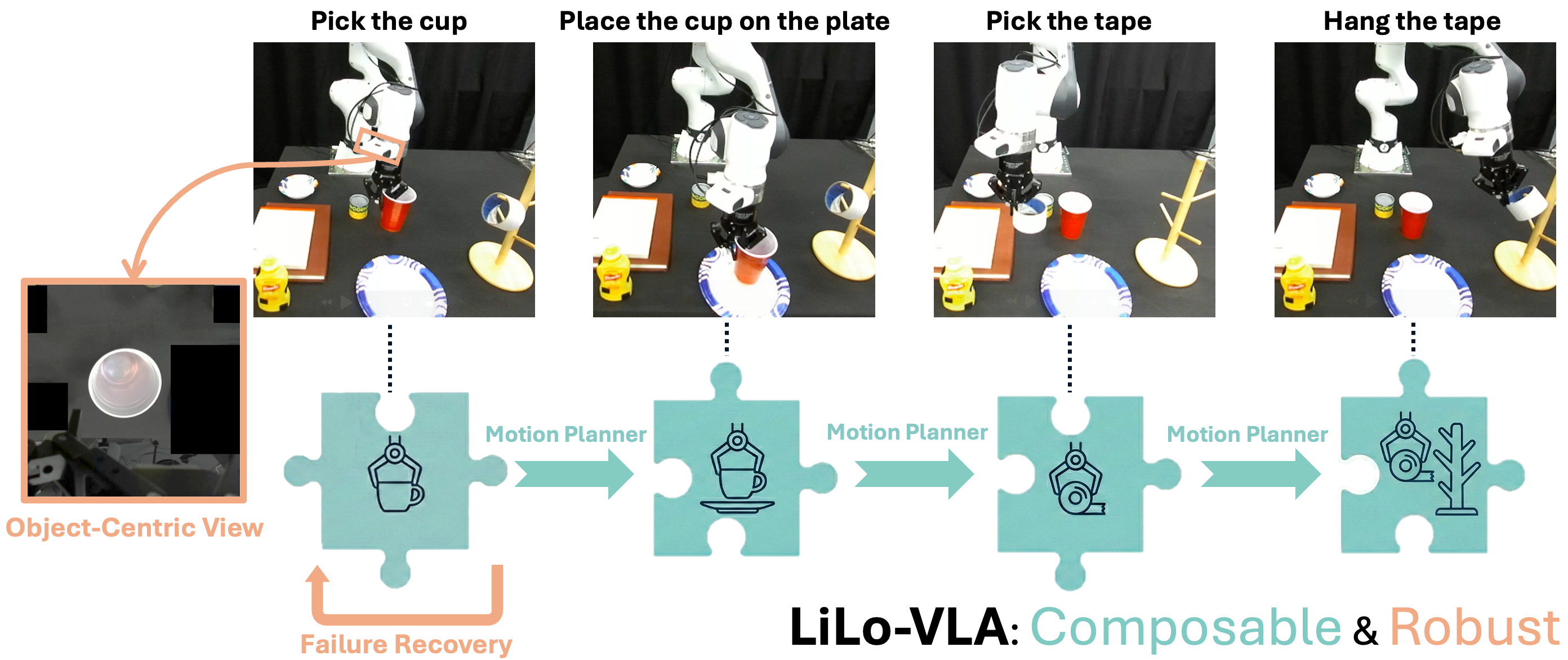}
    \vspace{-20pt}
    \captionof{figure}{\textbf{LiLo-VLA enables composable and robust manipulation.} LiLo-VLA solves long-horizon tasks by sequentially executing object-centric skill policies connected by robust motion planning. This enables zero-shot compositional generalization and robustness against cascading failures.}
\end{center}%
}]

\input{sections/abstract}

\IEEEpeerreviewmaketitle

\input{sections/introduction}
\input{sections/related_work}
\input{sections/methodology}

\input{sections/simu_experiments}
\input{sections/real_experiments}
\input{sections/conclusion}

\bibliographystyle{plainnat}
\bibliography{main}

\clearpage

\begin{center}
    {\Large \textbf{Appendix for LiLo-VLA}}
\end{center}

\setcounter{figure}{0}
\renewcommand{\thefigure}{S\arabic{figure}}
\setcounter{table}{0}
\renewcommand{\thetable}{S\arabic{table}}

\input{suppl_sections/overview}
\input{suppl_sections/additional_experimental_results}
\input{suppl_sections/benchmark_details}
\input{suppl_sections/system_design_details}

\input{suppl_sections/data_efficiency_and_scalability_analysis}
\input{suppl_sections/training_and_implementation_details}

\input{suppl_sections/infra_assets}
%% Use plainnat to work nicely with natbib. 

\end{document}

%% file: sections/abstract.tex
\begin{abstract}
    General-purpose robots must master long-horizon manipulation, defined as tasks involving multiple kinematic structure changes (e.g., attaching or detaching objects) in unstructured environments.
    While Vision-Language-Action (VLA) models offer the potential to master diverse atomic skills, they struggle with the combinatorial complexity of sequencing them and are prone to cascading failures due to environmental sensitivity.
    To address these challenges, we propose \textbf{LiLo-VLA} (\underline{Li}nked \underline{Lo}cal \underline{VLA}), a modular framework capable of zero-shot generalization to novel long-horizon tasks without ever being trained on them.
    Our approach decouples transport from interaction: a Reaching Module handles global motion, while an Interaction Module employs an object-centric VLA to process isolated objects of interest, ensuring robustness against irrelevant visual features and invariance to spatial configurations.  
    Crucially, this modularity facilitates robust failure recovery through dynamic replanning and skill reuse, effectively mitigating the cascading errors common in end-to-end approaches.
    We introduce a 21-task simulation benchmark consisting of two challenging suites: LIBERO-Long++ and Ultra-Long. In these simulations, \nm achieves a 69\% average success rate, outperforming Pi0.5 by 41\% and OpenVLA-OFT by 67\%. Furthermore, real-world evaluations across 8 long-horizon tasks demonstrate an average success rate of 85\%. Project page: \href{https://yy-gx.github.io/LiLo-VLA/}{https://yy-gx.github.io/LiLo-VLA/}.
\end{abstract}

%% file: sections/introduction.tex
% \begin{figure}[t]
%     \centering
%     \includegraphics[width=0.9\linewidth]{images/teaser_final.png} 
%     \caption{}
%     \label{fig:teaser}
% \end{figure}

\section{Introduction}

% \homanga{I may suggest making the intro slightly more general without specifically mentioning VLAs, system 1 system 2 etc...".... might be worth emphasizing that current generalist policies are limited to short-horizon activities, and the current long-horizon manipulation approaches are not generalizable requiring activity-specific heuristics! 
% .Long-Horizon Manipulation requiring multiple steps for completing an activity like cleaning, cooking, etc. remains a grand challenge in robotics...Recent approaches in generalizable policy learning are typically limited to single-staged manipulation -.... define long-hoirzon as requiring chaining multiple "atomic skills" like opening, closing, picking,...}
% The ultimate goal of general-purpose robotics is to master complex, long-horizon manipulation tasks in unstructured real-world environments.\gb{more precise definition of long-horizon somewhere in the intro or at least an example of the task?}
A grand goal of general-purpose robotics is to enable complex long-horizon manipulation, such as activities involving multiple skills (e.g., picking, placing, and pouring) to achieve high-level goals like cooking or cleaning.
While recent generalist policies have achieved impressive results on short-horizon, single-stage manipulation~\cite{chi2023diffusionpolicy,chi2024diffusionpolicy,Ze2024DP3,generalist2025gen0}, extending these capabilities to multi-stage activities remains a grand challenge.
Vision-Language-Action (VLA) models have emerged as a promising solution to bridge this gap, utilizing Internet-scale pre-training to enable diverse skill execution and semantic reasoning~\cite{kim2024openvla, kim2025fine, black2410pi0, intelligence2504pi0}.
However, applying current VLA paradigms to long-horizon tasks reveals two fundamental limitations.
% Vision-Language-Action (VLA) models have gained popularity as they offer the potential to master diverse skills and generalize to novel scenarios~\cite{kim2024openvla, kim2025fine, black2410pi0, intelligence2504pi0}.
% To bridge the gap between high-level semantic reasoning and low-level sensorimotor control, recent approaches typically adopt a ``System 1 + System 2'' cognitive architecture, where a Multimodal Large Language Models (MLLMs) acts as the high-level deliberative planner (System 2) and a reactive VLA policy serves as the low-level executor (System 1)~\cite{bjorck2025gr00t, shi2025hi}.
% However, current instantiations of this paradigm for long-horizon tasks suffer from two fundamental limitations.
% First, since existing methods typically learn from holistic long-horizon demonstrations, achieving compositional generalization, the ability to flexibly recombine skills, requires collecting data for every possible skill permutation, leading to a prohibitive combinatorial explosion~\cite{ross2011reduction, rajaraman2020toward}.\gb{This sentence is very long, cumbersome and difficult to understand. I also wouldn't frame it as these VLAs require collecting data for every possible skill. I would just say that these methods don't have compositional generalization and don't generalize to novel long-horizon tasks. It's a subtle but a different way to make a similar point, which to me sounds better}
First, existing methods lack compositional generalization: they struggle to adapt to novel task sequences that were not explicitly seen during training, as they fail to flexibly recombine learned atomic skills without extensive task-specific demonstrations~\cite{ross2011reduction, rajaraman2020toward}.
Second, long-horizon execution is prone to cascading failures: since VLA policies easily overfit to visual features or specific spatial configurations, they become brittle to minor variations, where a single failure at any stage jeopardizes the entire task sequence~\cite{yang2025boss, xu2025seeing}.

To mitigate the data requirements for compositional generalization, recent research has explored learning modular skills.
Approaches like PlanSeqLearn and local policy learning decouple tasks into atomic behaviors, employing motion planners for reaching and learned policies for interaction~\cite{dalal2024plan, dalal2025local}.
However, these methods struggle with skill chaining, often relying on labor-intensive reward engineering or ad-hoc local goals for motion planning integration that hinder scalability.
Alternatively, unified approaches like Long-VLA train a single end-to-end model to handle both transport and interaction phases via input masking~\cite{fan2025long}.
Yet, these methods remain data-inefficient as they couple global transport with interaction, forcing the network to learn geometric motion planning through unstructured data.
To address the second challenge of cascading failures, VLM-based recovery methods leverage the reasoning capabilities of foundation models to detect errors and trigger replanning~\cite{huang2022inner, liu2023reflect, guo2024doremi}.
However, these approaches primarily focus on task-level planning and rely on the critical assumption that underlying skills are robust.
Simply reinvoking a skill policy that is sensitive to irrelevant visual features is ineffective, as it often leads to repeated failures.
Moreover, execution failures frequently modify the spatial configuration of the scene, pushing the system outside the policy's initial state distribution.

To overcome these limitations, we introduce \textbf{\nm} (\textbf{Li}nked \textbf{Lo}cal \textbf{VLA}), a framework that synergizes the strengths of classical motion planning and VLA policies to master complex long-horizon manipulation. We adopt a modular control strategy that decouples execution into two distinct phases. A Reaching Module employs classical motion planners to navigate a robot's end-effector to a target vicinity, where control is handed over to an Interaction Module, an object-centric VLA dedicated to fine-grained atomic manipulation. Through this modular design and the seamless integration of learned policies with classical planners, \textbf{\nm} offers three decisive advantages. First, it achieves compositional generalization. By reusing object-centric atomic skills, the system can zero-shot generalize to novel long-horizon tasks without requiring any task-specific demonstration data. Second, it ensures robustness to environmental variations. By decomposing the scene and confining VLA observations strictly to the object of interest, the policy remains invariant to global workspace layouts and background clutter. Third, it enables inherent failure recovery. Rather than blindly retrying a failed skill, our system utilizes the Reaching Module to dynamically replan and reset the workspace, allowing for effective re-invocation of VLA skills. To comprehensively evaluate \textbf{\nm}, we curate a challenging benchmark and provide the corresponding dataset, consisting of 21 tasks across two distinct suites. Crucially, both suites enforce randomized skill sequencing to rigorously stress-test compositional generalization. The first enhances LIBERO-Long~\cite{liu2023libero} by introducing complex visual clutter to verify robustness against distractors. The second features custom ultra-long tasks chaining up to 16 atomic skills. This significantly exceeds the horizon length of prior benchmarks such as LIBERO-Long, which are typically limited to sequences of 3 to 4 skills, thereby challenging scalability. Our contributions in this work are four-fold:

% \begin{itemize}
%     \item We propose \nm, a modular framework for long-horizon tasks that enables zero-shot compositional generalization and achieves robustness to visual clutter and execution failures

%     \item We introduce a 21-task benchmark across two suites, ``LIBERO-Long++'' and ``Ultra-Long'', both of which evaluate compositional generalization while respectively focusing on visual robustness and extreme temporal scalability.
    
%     \item We conduct extensive simulation experiments showing that \nm achieves a 69\% average success rate, significantly outperforming SOTA VLA baselines such as Pi0.5~\cite{intelligence2504pi0} (28\%) and OpenVLA-OFT~\cite{kim2025fine} (2\%) across all benchmark suites.
    
%     \item We validate \nm on a real-world robotic system across 8 long-horizon tasks, featuring complex backgrounds and sequences of up to 8 skills. Our framework achieves an average success rate of 85\%.

% \end{itemize}

\begin{itemize}[leftmargin=*, nosep, itemsep=2pt]
    \item \textbf{LiLo-VLA Framework:} A modular framework for long-horizon tasks that enables zero-shot compositional generalization and achieves robustness to visual clutter and execution failures.
    
    \item \textbf{Evaluation Benchmarks:} A 21-task benchmark across two suites, ``LIBERO-Long++'' and ``Ultra-Long'', both of which evaluate compositional generalization while respectively focusing on visual robustness and extreme temporal scalability reaching up to 16 sequential skills.
    
    \item \textbf{Simulation Performance:} Extensive experiments show \nm achieves a 69\% average success rate, significantly outperforming Pi0.5 (28\%) and OpenVLA-OFT (2\%).
    
    \item \textbf{Real-World Validation:} Deployment on 8 long-horizon tasks (up to 8 skills) with complex backgrounds and varied sequence, achieving an 85\% average success rate.
\end{itemize}

%% file: sections/related_work.tex
\section{Related Work}
\label{sec:related_work}

\subsection{Vision-Language-Action Models}
Recent robot learning has been driven by VLA models that unify perception and control via Transformer architectures. Pioneering works like RT-2~\cite{zitkovich2023rt} and Octo~\cite{team2024octo} demonstrated the efficacy of large-scale pretraining. Building on this, OpenVLA~\cite{kim2024openvla} utilized LLM backbones to enhance performance, while OpenVLA-OFT~\cite{kim2025fine} improved control precision. Applications have expanded to bimanual control with RDT-1b~\cite{liu2024rdt} and humanoid embodiments via Project GR00t~\cite{bjorck2025gr00t}. Recently, Pi0~\cite{black2410pi0} and Pi0.5~\cite{intelligence2504pi0} leveraged flow matching and extensive real-world pretraining to capture complex action distributions. Despite improving atomic skill proficiency, these methods often overfit to visual signals and lack inherent compositional generalization.

\subsection{Long-Horizon Manipulation and Skill Chaining}
Solving long-horizon tasks requires effectively sequencing atomic behaviors to achieve a high-level goal. Classical Task and Motion Planning (TAMP) addresses this by combining symbolic search with geometric feasibility checks and engineered low-level controllers~\cite{kaelbling2011hierarchical, garrett2021integrated, silver2022learning}. However, TAMP typically assumes accurate state estimation and known object models, which limits its robustness and applicability in unstructured, real-world environments. More recently hierarchical learning approaches have integrated learned models for generating low-level motions~\cite{cheng2024nodtamp, cheng2023league, mishra2023generative,mandlekar2023human}. To reduce reliance on manually defined planning domains, researchers have explored using large language models to decompose abstract task instructions into intermediate, executable sub-goals. Frameworks, such as Code as Policy~\cite{liang2022code}, ProgPrompt~\cite{singh2022progprompt}, SayCan~\cite{ahn2022can} and VoxPoser~\cite{huang2023voxposer}, demonstrate strong semantic planning capabilities but often assume robust low-level execution. Most recently, LEAGUE~\cite{cheng2023league} and Plan-seq-learn~\cite{dalal2024plan} pioneered the integration of motion planning with RL trained policies for long-horizon tasks, while Long-VLA~\cite{fan2025long} introduced the first VLA architecture explicitly designed for long-horizon manipulation. However, the former relies on ad-hoc integration strategies that hinder scalability, while the latter is data-inefficient by coupling global transport with local interaction.

%% file: sections/methodology.tex
\section{Methodology}
\label{sec:methodology}

\begin{figure*}[t]
    \centering
    \includegraphics[width=\textwidth]{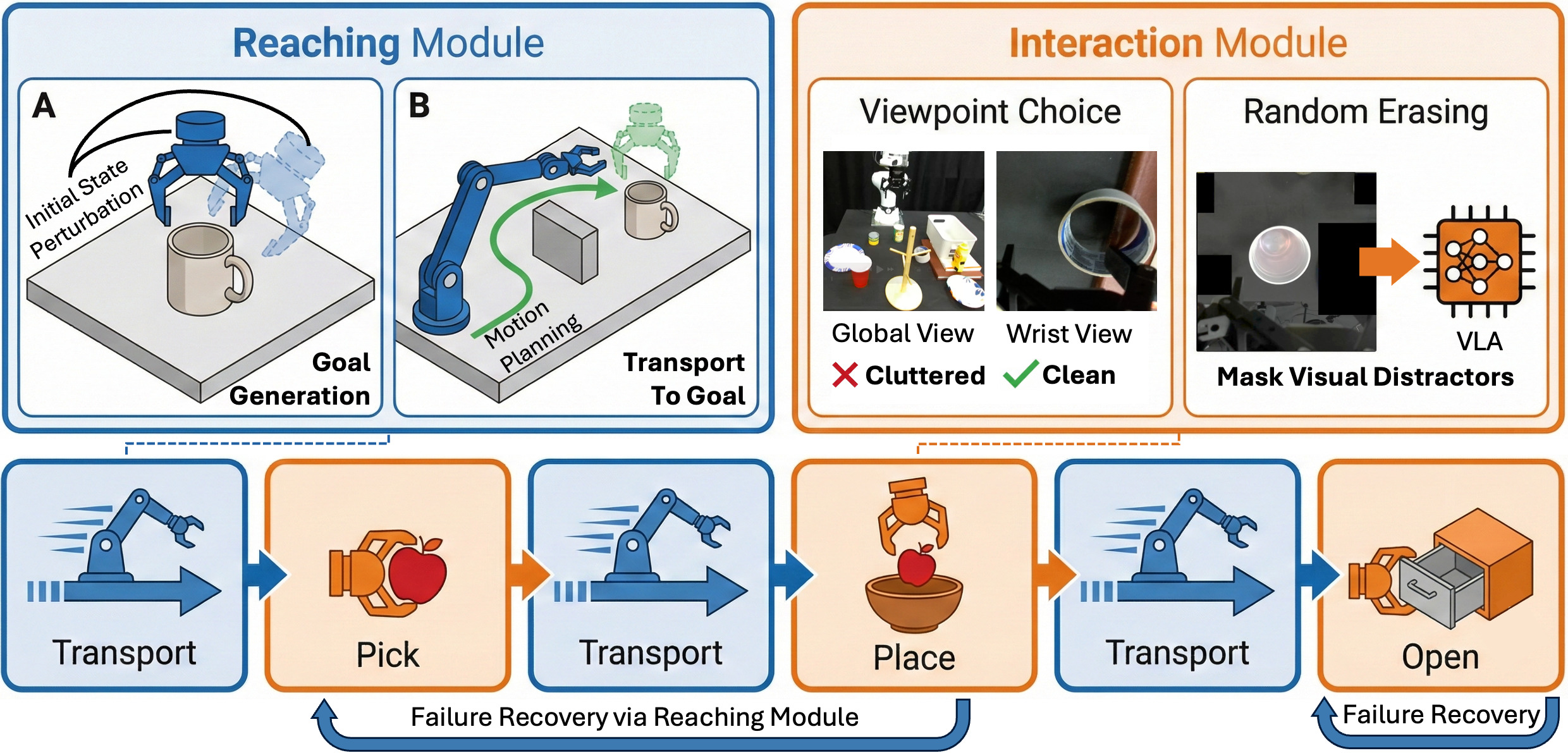}
    \vspace{-15pt}
    \caption{\textbf{Architecture of \nm.} Our framework decouples manipulation into two distinct phases. 
    \textbf{(Top Left) The Reaching Module} handles global transport via collision-free motion planning. It employs initial state perturbation during training to ensure the policy to be robust to pose errors during deployment. 
    \textbf{(Top Right) The Interaction Module} executes atomic skills via an object-centric VLA, strictly utilizing wrist-view observations and visual masking to eliminate environmental distractors.
    \textbf{(Bottom)} The system sequentially chains these modules, enabling closed-loop failure recovery where each skill's execution errors trigger a fallback to the Reaching Module for state resetting.}
    \label{fig:overview}
\end{figure*}

We describe \nm in four stages:
In Section~\ref{subsec:overview}, we first formalize the problem and outline the system architecture.
We then detail the Reaching Module, which handles global transport via robust motion planning, in Section~\ref{subsec:reaching_module}.
Subsequently, Section~\ref{subsec:interaction_module} introduces our Interaction Module, featuring the Object-Centric VLA designed for local robustness.
Finally, Section~\ref{subsec:execution_recovery} describes the closed-loop execution pipeline and our hierarchical failure recovery mechanism.

% Start with a short paragraph summarizing the section.
% "We present COVAL, a framework designed for... It consists of two core components: a Reaching Module for geometric transport and an Interaction Module for contact-rich manipulation. Finally, we describe the closed-loop execution and recovery strategy."

\subsection{Overview of \nm}
\label{subsec:overview}

% 1. Problem Formulation:
% Define a long-horizon task as a sequence of atomic skills T = {S_1, S_2, ..., S_N}.
% Explicitly state that we assume this sequence (and the Object of Interest for each skill) is provided by a high-level planner (e.g., a VLM or a human operator).
% Clarify that our focus is on the *robust execution* of this plan, not the planning itself.

% 2. Architecture Diagram Reference:
% "As shown in Figure 2..." 
% Briefly explain the flow: The system iterates through the skills. For each skill, the Reaching Module first transports the end-effector to a robust starting pose, followed by the Interaction Module which executes the learned policy.

% 3. Core Philosophy:
% Mention the decoupling strategy again: using Motion Planning (MP) for what it's good at (collision-free, long-distance) and VLA for what it's good at (local interaction), ensuring compositional generalization.

Formally, we define a long-horizon manipulation task as a sequence of symbolic actions $\mathcal{T} = \{a_1, a_2, \dots, a_N\}$. 
Each action $a_i$ is formulated as a parameterized predicate $\alpha_i(o_i, \rho_i)$, where $\alpha_i$ denotes the primitive skill category (e.g., \texttt{Pick}, \texttt{Place}), $o_i$ represents the reference object that establishes the local coordinate frame for execution (e.g., the target object for a Pick action, or the receptacle for a Place action), and $\rho_i$ encapsulates auxiliary parameters such as constraints. 
% In this work, we operate under the assumption that this symbolic plan $\mathcal{T}$ is provided by an external high-level planner, such as a MLLM or a PDDL-based solver. 
Although such high-level task skeleton can be generated automatically---either by symbolic planners~\cite{silver2021learning,garrett2021integrated,liang2022search} or by foundation models (LLMs/VLMs)~\cite{lin2023text2motion,feng2025reflective}---our focus in this work is on generating low-level robot motions that faithfully realize a given task plan $\mathcal{T}$ and achieve the desired goal under realistic geometric and dynamical constraints.

% As illustrated in Figure~\ref{fig:overview}, \nm adopts a modular architecture to ground and execute these symbolic actions sequentially.
% For each action $a_i$, the system executes the corresponding atomic skill through a two-phase process: 
% first, the Reaching Module transports the end-effector from its current configuration to a robust approach pose defined relative to $o_i$; 
% subsequently, the Interaction Module is activated to perform the contact-rich manipulation using a learned policy centered on $o_i$.
% This design philosophy effectively decouples the problem space: it utilizes motion planning for long-distance, collision-free transport and leverages the Object-Centric VLA for local interaction, ensuring that the expensive learning capacity is focused exclusively on mastering the diverse dynamics of atomic skills. \textcolor{red}{TO BE MODIFIED after finishing Fig 2}.

As illustrated in Figure~\ref{fig:overview}, \nm adopts a modular architecture to ground and execute these symbolic actions sequentially. For each action $a_i$, the system executes the corresponding atomic skill through a two-phase process. First, the Reaching Module (Figure~\ref{fig:overview} Top-Left) transports the end-effector from its current configuration to a robust approach pose defined relative to $o_i$ via motion planning. Subsequently, the Interaction Module (Figure~\ref{fig:overview} Top-Right) is activated to perform the contact-rich manipulation using a learned policy centered on $o_i$ with visual masking. As depicted in the bottom pipeline of Figure~\ref{fig:overview}, this design philosophy effectively decouples the problem space. It utilizes motion planning for collision-free transport and leverages the Object-Centric VLA for local interaction, ensuring that the expensive learning capacity is focused exclusively on mastering the diverse dynamics of atomic skills while enabling autonomous failure recovery.

\subsection{The Reaching Module: Global Transport}
\label{subsec:reaching_module}

The Reaching Module serves as the bridge between distinct atomic skills, addressing the challenge of navigating the end-effector from the termination state of the previous skill to the initial state of the next. By leveraging a motion planner, this module ensures collision avoidance over long distances and guides the robot to a precise approach pose suitable for the downstream object-centric VLA. This approach pose also serves as a deterministic target for the motion planner during deployment.

\subsubsection{Relative Goal Generation with Perturbation}
\label{subsubsec:goal_generation}

The core function of this module is to determine the robust \textit{Approach Pose}, $T_{approach} \in SE(3)$, which serves as the deterministic target for the motion planner. 
We derive this pose via a relative transformation from the reference object's frame: $T_{approach} = T_{o_i} \cdot T_{offset}(\alpha_i)$.
Here, the offset transformation $T_{offset}(\alpha_i)$ is constructed to enforce a fixed face-down orientation relative to the object surface for all skills, while applying a skill-specific translation vector defined by $\alpha_i$ (e.g., a vertical clearance $h_{pick}$ for picking tasks).

To ensure the downstream VLA policy is resilient to noise stemming from motion planning and perception, we introduce a perturbation strategy. 
% During the demonstration generation phase, instead of initializing demonstrations strictly from the canonical $T_{approach}$, we sample the initial state of every training demonstration, $T_{init}$, from a noisy distribution centered on the approach pose: $T_{init} = T_{approach} \cdot \Delta T, \Delta T = exp(\hat{\xi})$, where $\xi \sim \mathcal{N}(0, \Sigma)$ represents random perturbations in both translation and rotation.
During demonstration generation, rather than always initializing from the canonical approach pose $T_{{approach}}$, we randomize the start pose of each training trajectory by sampling
\[
T_{{init}} = T_{{approach}}\,\Delta T,\qquad \Delta T=\exp(\hat{\xi}),
\]
where $\xi \sim \mathcal{N}(0,\Sigma)$ defines a zero-mean perturbation in $\mathrm{SE}(3)$, injecting noise in both translation and rotation. The robot is transported to this perturbed state $T_{init}$ to initiate each data collection episode.
Consequently, during the deployment phase, while the Reaching Module targets the canonical $T_{approach}$, the VLA policy acquires strong local generalization capabilities by training on the dispersed distribution of $T_{init}$. This allows the Interaction Module to effectively compensate for deviations caused by pose estimation errors and imperfect motion planning convergence. 
% \gb{This paragraph is very difficult for me to follow. Many variables and no high-level motivation/intuition}

\subsubsection{Collision-Free Motion Planning}
\label{subsubsec:motion_planning}

Given the determined target pose, $T_{approach}$, during deployment or $T_{init}$ during demonstration generation, we employ MPLib~\cite{guomplib} to generate a collision-free trajectory. 
This planner integrates the environmental point cloud with the robot's kinematic chain to solve for a feasible path, ensuring safe global transport to the interaction start state without requiring any learning.

\subsection{The Interaction Module: Object-Centric VLA}
\label{subsec:interaction_module}
The reaching module brings the robot end-effector into proximity with the object. We model robot–object interaction by learning an end-to-end visuomotor policy $\pi_\theta$ that reactively predicts end-effector actions from the current observations. To ensure robustness against the spatial variability inherent in long-horizon tasks, we introduce a strictly object-centric design.

\subsubsection{Object-Centric Observation Space}
\label{subsubsec:obs_space}

% The observation space consists of two primary components: visual input and proprioceptive state.

% For visual input, we exclusively utilize the egocentric RGB images captured by the wrist-mounted camera. 

To ensure a strictly object-centric representation, our observation space relies exclusively on egocentric visual inputs captured by the wrist-mounted camera. We deliberately exclude static third-person views to mitigate the observation space shift (OSS) problem common in long-horizon manipulation~\cite{yang2025boss}, defined as the performance degradation caused by task-irrelevant visual changes across different stages of a long-horizon task. Since the robot base or global position may vary between skill executions, relying on fixed global cameras can introduce inconsistent visual features. In contrast, the wrist view maintains a consistent perspective relative to the workspace during interaction. This observation locality helps the robot focus on task-relevant objects, reducing the negative impact of visual distractors. We provide empirical validation for this design choice in Section~\ref{subsubsec:ablation_vision}, demonstrating that a wrist-only policy outperforms multi-view configurations in atomic skill success rates.

% suppl
% For proprioception, we discard global coordinates in favor of relative spatial information. Specifically, we define the proprioceptive state as the 6-DoF pose of the end-effector relative to the local coordinate frame of the reference object $o_i$. Let $T_{world}^{ee}$ denote the global end-effector pose and $T_{world}^{obj}$ denote the global pose of the reference object. The input to the policy is computed as $T_{obj}^{ee} = (T_{world}^{obj})^{-1} T_{world}^{ee}$. This formulation renders the policy invariant to the global position and orientation of the workspace. Consequently, the policy learns the relative interaction dynamics rather than memorizing absolute spatial locations, allowing zero-shot generalization to novel workspace configurations.

\subsubsection{Visual Clutter Augmentation}
\label{subsubsec:visual_augmentation}

VLA policies frequently overfit to environmental distractors, such as objects relevant to other skills~\cite{xu2025seeing}. To eliminate this interference during deployment, we apply a heuristic masking strategy that covers non-target objects with black rectangles derived from their bounding boxes. However, these artificial occlusions constitute a significant visual domain shift, potentially leading to out-of-distribution (OOD) failures for a standard policy.

To ensure robustness against these artifacts, we introduce a mask-aware data augmentation strategy during training, inspired by \cite{zhong2020random}. We utilize segmentation to partition each training frame into foreground (the gripper, the reference object, and any grasped object) and background (all other pixels). We then apply a random erasing augmentation that overlays black rectangles exclusively onto the background regions. The number of rectangles and their total area ratio are sampled uniformly from a predefined range. This training procedure effectively simulates the visual artifacts introduced by our deployment masking, ensuring that the masked inference observations remain within the learned distribution of the policy.

\subsection{Compositional Execution and Failure Recovery}
\label{subsec:execution_recovery}

The complete inference pipeline integrates the proposed modules into a cohesive control loop, as summarized in Algorithm~\ref{alg:inference_loop}.
% \gb{Instead of an algorithm it would be good to have a high-level figure that illustrates this whole complex pipeline}

% --- ALGORITHM BLOCK START ---
\begin{algorithm}[t]
\caption{Compositional Inference with Closed-Loop Recovery}
\label{alg:inference_loop}
\begin{algorithmic}[1]
\Require Skill sequence $\mathcal{T} = \{a_1, a_2, \dots, a_N\}$
\Require Modules: Reaching $\mathcal{M}_{\text{reach}}$, Interaction $\mathcal{M}_{\text{int}}$

\State Initialize skill index $i \leftarrow 1$
\While{$i \leq N$}
    \State $p_{\text{target}} \leftarrow \text{EstimateObjectPose}(a_i)$
    \State $\mathcal{M}_{\text{reach}}.\text{MoveToAbove}(p_{\text{target}})$
    \State $\mathcal{M}_{\text{int}}.\text{ExecuteSkill}(a_i)$
    \State $success \leftarrow \text{VerifyCondition}(a_i)$
    
    \If{success}
        \State $i \leftarrow i + 1$
    \Else
        \If{$a_i$ involves holding object}
            \State $i \leftarrow \text{LastPickIndex}(\mathcal{T}, i)$
        \Else
            \State \textbf{continue}
        \EndIf
    \EndIf
\EndWhile
\end{algorithmic}
\end{algorithm}
% --- ALGORITHM BLOCK END ---

\subsubsection{Sequential Execution Pipeline}
\label{subsubsec:pipeline}

For each atomic skill $a_i$ in the plan, the system orchestrates the execution through a standardized pipeline (Alg.~\ref{alg:inference_loop}, Lines 3-6). 
First, the system estimates the 6D pose of the target object relevant to the current skill. 
This pose is ingested by the Reaching Module ($\mathcal{M}_{\text{reach}}$), which plans and executes a collision-free trajectory to navigate the end-effector to the approach pose, denoted as $T_{\text{approach}}$. This alignment ensures a consistent geometric initialization for the subsequent manipulation.
Subsequently, the Interaction Module ($\mathcal{M}_{\text{int}}$) takes control to perform the fine-grained manipulation task using the VLA policy. 
Finally, the execution concludes with a geometric verification function $\mathcal{V}(a_i)$, which evaluates the spatial configuration of the object of interest to determine if the skill's effect has been satisfied.
% \gb{You have a lot of variables. Do you need all of them? It significantly increases burden on the reader to keep track of all of them}

\subsubsection{Closed-Loop Recovery Mechanism}
\label{subsubsec:recovery}

The modular architecture enables robust recovery from execution failures by adapting the control flow based on the semantic state of the robot (Alg.~\ref{alg:inference_loop}, Lines 8-12).
When a failure is detected by $\mathcal{V}(a_i)$ for a skill that does not involve holding an object (e.g., a \textit{Pick} attempt), the system triggers a local retry. By maintaining the current skill index $i$ and re-initiating the loop, the system forces a re-evaluation of the spatial state. Specifically, this updates the object pose and resets the end-effector to $T_{\text{approach}}$, correcting any transient errors caused by the failed interaction and ensuring the policy is conditioned on the latest observation.
In contrast, if a failure occurs during a skill that requires object retention (e.g., a \textit{Place} operation), we conservatively assume that the object has been dropped or lost during transport. Operating under this assumption, a local retry is deemed risky. Therefore, the system backtracks the index $i$ to the most recent \textit{Pick} skill, ensuring the robot re-acquires the object before attempting the transport task again.

%% file: sections/simu_experiments.tex
\section{Simulation Experiments}
\label{sec:experiments}

\subsection{Experimental Setup}
\label{subsec:setup}

\begin{figure*}[t]
    \centering
    \includegraphics[width=\linewidth]{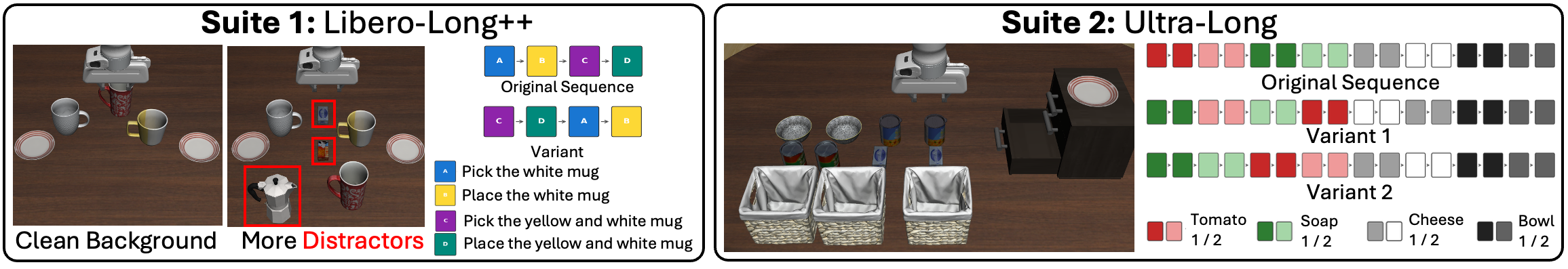} 
    \caption{\textbf{Overview of Evaluation Benchmarks.} We introduce two suites to evaluate long-horizon manipulation: Suite 1 (LIBERO-Long++) focuses on visual robustness by introducing more complex backgrounds with multiple distractors (highlighted in red), while Suite 2 (Ultra-Long) tests temporal scalability with task sequences extending up to 16 steps. Both suites incorporate multiple variant configurations with permuted skill orders to rigorously assess zero-shot compositional generalization.}
    \label{fig:bm_illustration}
\end{figure*}

\subsubsection{Benchmarks}
\label{subsubsec:benchmarks}

To evaluate zero-shot compositional generalization and robustness to cascading failures, we curate a 21-task benchmark across two distinct suites within the LIBERO environment~\cite{liu2023libero}, derived from 9 core scenarios.

\textbf{Suite 1: LIBERO-Long++.} We select 6 tasks from LIBERO-Long that are amenable to sequence re-ordering and augment them with randomized visual distractors (e.g., mugs, cans). As shown in Fig.~\ref{fig:bm_illustration}, this setup increases visual complexity to test robustness, forcing the policy to filter background clutter and attend strictly to task-relevant objects. 

\textbf{Suite 2: Ultra-Long.} We design 3 tasks with increasing complexity: Kitchen Organization (9 steps), Cooking Preparation (10 steps), and Living Room Organization (16 steps). As illustrated in Fig.~\ref{fig:bm_illustration}, the 16-step task presents a extreme challenge in workspace saturation; here, the high density of objects and receptacles pushes the kinematic reachability limits of a fixed-base tabletop robot. This suite systematically evaluates the system's ability to maintain coherent execution and temporal scalability over extended horizons.

\textbf{Evaluation Protocol.} To assess true compositionality rather than trajectory memorization, we evaluate multiple skill permutations for each scenario: 2 permutations for each of the 6 ``Suite 1'' scenarios and 3 for the 3 ``Suite 2'' scenarios (21 tasks in total). This protocol rigorously tests the system’s ability to ground novel skill execution orders in zero-shot way.

\subsubsection{Baselines}
\label{subsubsec:baselines}

We compare \nm against two state-of-the-art generalist policies: \textbf{Pi 0.5}~\cite{intelligence2504pi0}, a flow-matching VLA, and \textbf{OpenVLA-OFT}~\cite{kim2025fine}, a highly optimized version of  OpenVLA~\cite{kim2024openvla}.

\textbf{Setup for Suite 1.} For standard sequences, baselines are LoRA fine-tuned~\cite{hu2022lora} on the original continuous long-horizon demonstrations to learn skill transitions in a fixed order. For variant sequences with novel orders, we evaluate zero-shot performance by prompting the models with re-ordered language instructions.

\textbf{Setup for Suite 2.} Since collecting continuous demonstrations for every possible sequence permutation is combinatorially intractable, we train baselines on the aggregate dataset of constituent atomic skills to evaluate their inherent compositionality. During inference, we employ language chaining, sequentially prompting the model with sub-task descriptions to execute skills one-by-one. This evaluates whether VLAs can compose learned behaviors into novel long-horizon chains without requiring task-specific trajectory data.

\subsubsection{Evaluation Metrics}
\label{subsubsec:metrics}

We evaluate each task configuration over 10 trials and report performance using two standard metrics. \textbf{Success Rate (SR)} measures the percentage of episodes that reach the final goal state. Crucially, we enforce a strict standard: we count an episode as a success only if the robot executes the entire sequence of skills in the exact required order. \textbf{Average Progress (AP)} calculates the ratio of completed skills starting from the first step. We count only the continuous sequence of correct actions; if a skill is performed out of order, we stop counting immediately.

\subsubsection{Implementation Details}
\label{subsubsec:details}

To isolate manipulation capability from perception noise, we utilize simulator-provided ground-truth poses and segmentation masks. This oracle assumption ensures reported failures are attributable solely to the manipulation policy or motion planning rather than upstream estimation errors.

\textbf{Data Generation.} We segment LIBERO-90 demonstrations into atomic skills and augment them via MPLib to bridge the planner-policy domain gap. By generating perturbed initial states around the approach pose and planning trajectories back to the original demonstrations, we create a dataset of skills initialized from diverse starting configurations.

\textbf{Policy Architecture.} We adopt OpenVLA-OFT~\cite{kim2025fine} as our backbone, retaining original hyperparameters for controlled comparison. To enforce object-centricity, we utilize the wrist camera view and apply random erasing to background pixels, mitigating visual overfitting to distractors.

\textbf{Deployment.} During inference, the system executes the symbolic plan sequentially. For each skill, the Reaching Module plans a collision-free path to an approach pose defined by a fixed offset; the Interaction Module then takes control using language instructions. We implement the failure detector $\mathcal{V}(S_i)$ via geometric heuristics to trigger closed-loop recovery.

\subsection{Main Results: Zero-Shot Compositionality and Scalability}
\label{subsec:main_results}

\input{tables/main_table}

The quantitative results in Table \ref{tab:system_results} demonstrate that \nm significantly outperforms state-of-the-art baselines across all metrics achieving an average success rate of 69\% compared to 28\% for Pi0.5~\cite{intelligence2504pi0} and 2\% for OpenVLA-OFT~\cite{kim2025fine}. This substantial performance gap highlights the limitations of monolithic policies in handling the complexity of long-horizon manipulation. In the LIBERO-Long++ suite the results reveal a critical weakness in current VLAs where Pi0.5 achieves a high success rate on Original sequences (83\%) which is statistically comparable to \nm (78\%, $p > 0.05$) yet collapses to 0\% on Variant sequences. We observe that Pi0.5 frequently ignores the altered language instructions in these variant tasks and persists in executing the sequence order seen during training which confirms that the model overfits to the demonstrated trajectories rather than grounding the current language command. In contrast \nm maintains robust performance (85\%) by effectively isolating atomic skills. Furthermore in the ultra-long tasks of Suite 2 both baselines fail completely with a 0\% success rate while \nm maintains a 44\% success rate. We attribute this failure to the coupling of global transport and local interaction in baseline models. Long-horizon tasks inherently involve evolving geometric configurations and spatial distribution shifts. Consequently baselines struggle to generalize to these changing layouts without intractable amounts of transition data. By explicitly decoupling global transport via a motion planner \nm renders the interaction policy invariant to these global spatial shifts and effectively addresses the scalability bottleneck.

\subsection{Ablation Studies: Dissecting the \nm Framework}
\label{subsec:ablations}
% LOGIC: General introduction. State that we validate the design choices of the three core modules: Reaching, Interaction, and Recovery.
% We conduct a series of ablation studies to dissect the critical design choices of the \nm framework. Section \ref{subsubsec:ablation_reaching} validates the necessity of the Reaching Module and the robustness of the interface against initialization noise. Section \ref{subsubsec:ablation_vision} analyzes the impact of the object-centric observation space and visual augmentation within the Interaction Module. Finally, Section \ref{subsubsec:ablation_recovery} demonstrates the role of the closed-loop recovery mechanism in ensuring scalability for ultra-long tasks.

\subsubsection{Necessity and Robustness of the Reaching Module}
\label{subsubsec:ablation_reaching}

We first investigate the fundamental necessity of decoupling transport from interaction by removing the Reaching Module as shown in Table \ref{tab:system_results}. To address potential concerns regarding fair comparison and strictly isolate the impact of the modular architecture, we provided this ablation baseline with the same privileged ground-truth object poses used by our full method. Despite having access to this oracle geometric information, the policy fails completely with a 0\% success rate across all tasks. This failure confirms that standard VLA architectures cannot implicitly learn long-horizon transport dynamics from atomic skill demonstrations alone even when the target location is known. Consequently, an explicit motion planner is not merely an auxiliary component but a structural prerequisite for bridging spatially distributed skills.

Beyond necessity, we analyze the robustness of the interface between the motion planner and the policy. In real-world deployment, the system must contend with inevitable inaccuracies stemming from perception noise and imperfect motion planning convergence. To evaluate resilience to these factors, we conduct a comprehensive evaluation aggregating all 36 atomic skills from both Suite 1 and Suite 2, introducing execution noise to the initial poses of evaluation episodes as shown in Fig. \ref{fig:ablation_perturbation}. A baseline policy trained solely on canonical trajectory endpoints performs adequately under perfect initialization but degrades significantly under this noisy execution setting. In contrast, our full model which incorporates initial state perturbation during training maintains robust performance with no statistically significant drop in success rate ($p > 0.05$). This stability confirms that our data generation strategy effectively bridges the domain gap between the precision of the motion planner and the local generalization required by the policy.

\input{tables/perturbation}
% SUGGESTION: Ensure the caption for this figure (defined inside the input file) emphasizes: "Robustness to Motion Planning Noise and Global Distribution Shifts."

\subsubsection{Visual Robustness via Object-Centric Design}
\label{subsubsec:ablation_vision}

\begin{figure}[t]
    \centering
    \includegraphics[width=0.9\linewidth]{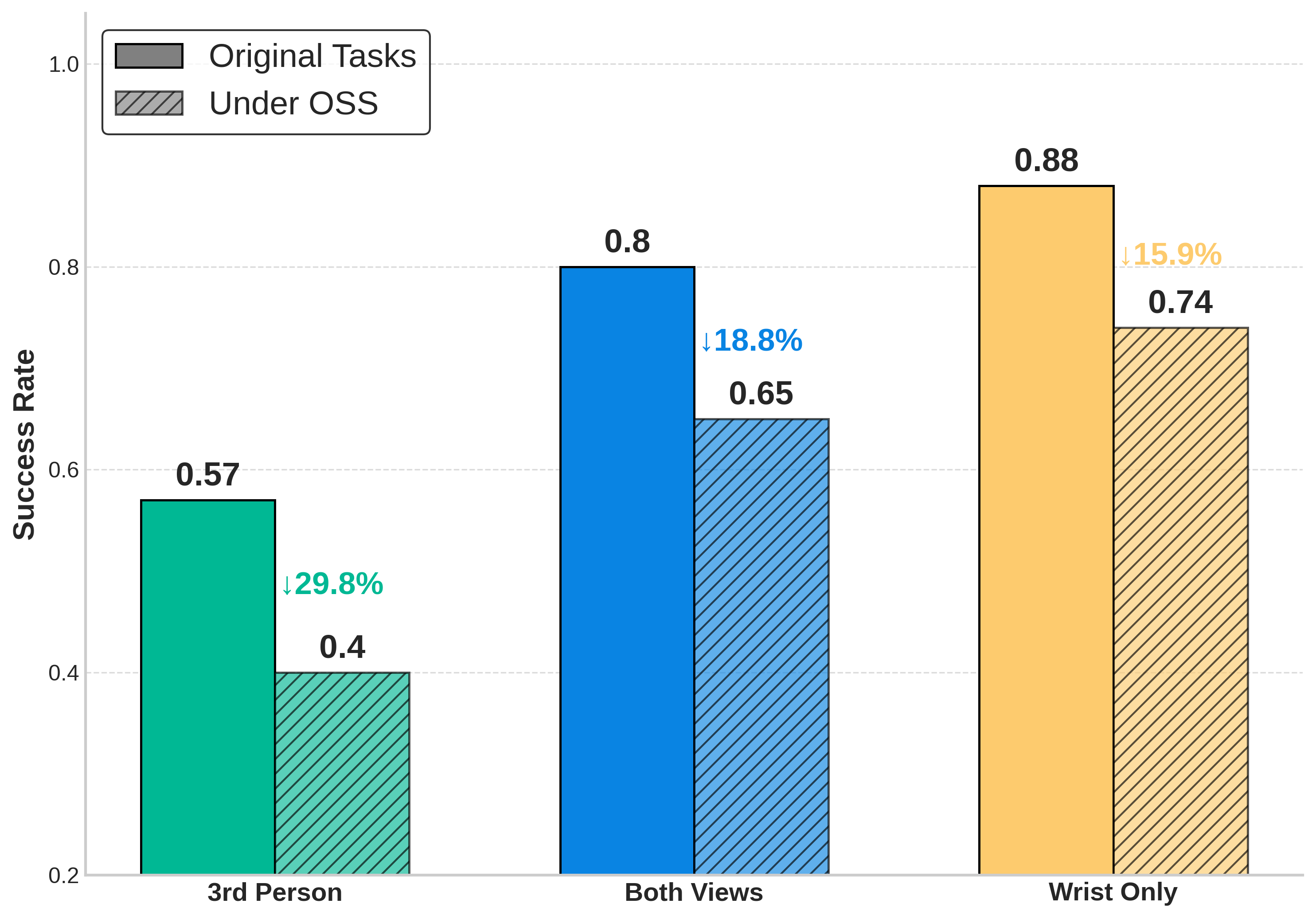} 
    \caption{\textbf{Comparison across different camera configurations.} Our wrist-only design achieves the highest success rate with the minimal performance drop under OSS.}
    \label{fig:ablation_viewpoint}
\end{figure}

Next, we validate the design choices of our interaction module, focusing on viewpoint selection and visual augmentation. To evaluate resilience to visual disturbances, we conduct experiments on the BOSS-C1 benchmark~\cite{yang2025boss}, which stress-tests policies against Observation Space Shift, defined as the phenomenon where changes in task-irrelevant visual predicates within the observation space disrupt the performance of a learned policy. As shown in Fig. \ref{fig:ablation_viewpoint}, the ``Wrist Only'' configuration achieves the highest absolute success rate on original tasks (0.88). We attribute this to the wrist-mounted perspective, where the target object occupies a larger portion of the field of view, facilitating feature extraction compared to global views. Crucially, under OSS conditions, the ``Wrist Only'' configuration also exhibits the lowest Ratio Performance Delta (15.9\%) compared to ``Both Views'' (18.8\%) and ``3rd Person'' (29.8\%). This confirms that the wrist view is inherently more robust to background clutter and task-irrelevant changes in the environment.

Complementing the viewpoint choice, the ``w/o Masking'' row in Table \ref{tab:system_results} highlights the necessity of our visual augmentation. Removing the random erasing strategy leads to a sharp decline in the overall success rate from 69\% to 48\%. This significant drop indicates that even with a wrist camera, the policy's receptive field inevitably includes environmental distractors. Explicitly masking these irrelevant regions during training is therefore essential to strictly enforce object-centricity and prevent the policy from overfitting to spurious visual features.

% \subsubsection{Efficacy of Closed-Loop Recovery}
% \label{subsubsec:ablation_recovery}
% % LOGIC: This section validates the "Scalability" claim.
% % 1. Reference Table 1 'w/o Recovery'.
% % 2. Contrast Suite 1 vs. Suite 2.
% %    - Suite 1 (Short): Performance gap is smaller (Recovery is helpful but not critical).
% %    - Suite 2 (Ultra-Long): Performance gap is massive (0% without recovery vs. 44% with recovery).
% % 3. Insight: In long chains (16 steps), the probability of error approaches 1.0. Recovery is not just an optimization but a strict prerequisite for scalability.

\subsubsection{Efficacy of Closed-Loop Recovery}
\label{subsubsec:ablation_recovery}

Finally, we examine the role of our closed-loop recovery mechanism. The ``w/o Recovery'' row in Table \ref{tab:system_results} reveals that removing this module leads to a catastrophic performance drop, with the overall average success rate plummeting from 69\% to 8\%. This failure is particularly absolute in the ultra-long tasks of Suite 2, where the success rate collapses to 0\% without recovery compared to 44\% with the full system. This dramatic difference demonstrates that closed-loop recovery is not merely an optimization but a structural prerequisite for scaling to extended horizons, where open-loop execution inevitably fails due to cumulative errors.

%% file: tables/main_table.tex
% --- Ensure these are in your preamble ---
% \usepackage{booktabs}
% \usepackage{multirow}
% \usepackage{graphicx}
% \usepackage[table]{xcolor}
% \definecolor{mygray}{gray}{0.9}
% ---------------------------------------

\begin{table*}[t]
\centering
\caption{\textbf{Evaluation on Libero-Long++ and Ultra-Long.} We report Success Rate (SR, \%) and Average Progress (AP). \textbf{Suite 1} tests robustness to visual clutter. \textbf{Suite 2} tests scalability (up to 16 skills). \colorbox{mygray}{Gray columns} indicate average performance.}
% \gb{I feel like these comparisons to OpenVLA and Pi0.5 are extremely unfair and actually very misleading. Your model assumes ground truth object position, which is a significant advantage, whereas the other VLA baselines don't. You never acknowledge that the comparison is not fair since your method assumes more information. If you used an off-the-shelf object detector for these experiments (maybe you do but I just missed the details?  then the comparison is ok.}
% \gb{Also, i feel like you should have ablations in a separate table}
\label{tab:system_results}
\setlength{\tabcolsep}{8pt} % Adjusted slightly to fit the extra columns better
\resizebox{\textwidth}{!}{%
% Updated column definition: Suite 2 now has 8 columns (cccccccc)
\begin{tabular}{l|cccccc|cccccccc|cc}
\toprule
\multirow{3}{*}{\textbf{Method}} & \multicolumn{6}{c|}{\textbf{Suite 1: Visual Clutter (LIBERO-Long++)}} & \multicolumn{8}{c|}{\textbf{Suite 2: Scalability (Ultra-Long)}} & \multicolumn{2}{c}{\textbf{Overall}} \\
\cmidrule(lr){2-7} \cmidrule(lr){8-15} \cmidrule(lr){16-17}
 & \multicolumn{2}{c}{\textit{Original}} & \multicolumn{2}{c}{\textit{Variant}} & \multicolumn{2}{c|}{\cellcolor{mygray}\textit{Avg.}} & \multicolumn{2}{c}{\textit{Original}} & \multicolumn{2}{c}{\textit{Variant 1}} & \multicolumn{2}{c}{\textit{Variant 2}} & \multicolumn{2}{c|}{\cellcolor{mygray}\textit{Avg.}} & \multicolumn{2}{c}{\cellcolor{mygray}\textbf{Avg.}} \\
\cmidrule(lr){2-3} \cmidrule(lr){4-5} \cmidrule(lr){6-7} \cmidrule(lr){8-9} \cmidrule(lr){10-11} \cmidrule(lr){12-13} \cmidrule(lr){14-15}
 & \textbf{SR} & \textbf{AP} & \textbf{SR} & \textbf{AP} & \multicolumn{1}{c}{\cellcolor{mygray}\textbf{SR}} & \multicolumn{1}{c|}{\cellcolor{mygray}\textbf{AP}} & \textbf{SR} & \textbf{AP} & \textbf{SR} & \textbf{AP} & \textbf{SR} & \textbf{AP} & \multicolumn{1}{c}{\cellcolor{mygray}\textbf{SR}} & \multicolumn{1}{c|}{\cellcolor{mygray}\textbf{AP}} & \multicolumn{1}{c}{\cellcolor{mygray}\textbf{SR}} & \multicolumn{1}{c}{\cellcolor{mygray}\textbf{AP}} \\
\midrule
\multicolumn{17}{l}{\textit{Baselines}} \\
\addlinespace[2pt]
Pi0.5 & \textbf{83\%} & \textbf{93\%} & 0\% & 0\% & \multicolumn{1}{c}{\cellcolor{mygray}42\%} & \multicolumn{1}{c|}{\cellcolor{mygray}46\%} & 0\% & 1\% & 0\% & 0\% & 0\% & 0\% & \multicolumn{1}{c}{\cellcolor{mygray}0\%} & \multicolumn{1}{c|}{\cellcolor{mygray}0.3\%} & \multicolumn{1}{c}{\cellcolor{mygray}28\%} & \multicolumn{1}{c}{\cellcolor{mygray}31\%} \\
OpenVLA-OFT & 7\% & 11\% & 0\% & 0\% & \multicolumn{1}{c}{\cellcolor{mygray}3\%} & \multicolumn{1}{c|}{\cellcolor{mygray}5\%} & 0\% & 0\% & 0\% & 0\% & 0\% & 0\% & \multicolumn{1}{c}{\cellcolor{mygray}0\%} & \multicolumn{1}{c|}{\cellcolor{mygray}0\%} & \multicolumn{1}{c}{\cellcolor{mygray}2\%} & \multicolumn{1}{c}{\cellcolor{mygray}4\%} \\
\midrule
\multicolumn{17}{l}{\textit{Ablations}} \\
\addlinespace[2pt]
w/o Reaching & 0\% & 0\% & 0\% & 0\% & \multicolumn{1}{c}{\cellcolor{mygray}0\%} & \multicolumn{1}{c|}{\cellcolor{mygray}0\%} & 0\% & 0\% & 0\% & 0\% & 0\% & 0\% & \multicolumn{1}{c}{\cellcolor{mygray}0\%} & \multicolumn{1}{c|}{\cellcolor{mygray}0\%} & \multicolumn{1}{c}{\cellcolor{mygray}0\%} & \multicolumn{1}{c}{\cellcolor{mygray}0\%} \\
w/o Masking & 67\% & 80\% & 77\% & 87\% & \multicolumn{1}{c}{\cellcolor{mygray}72\%} & \multicolumn{1}{c|}{\cellcolor{mygray}83\%} & 0\% & 16\% & 0\% & 32\% & 0\% & 10\% & \multicolumn{1}{c}{\cellcolor{mygray}0\%} & \multicolumn{1}{c|}{\cellcolor{mygray}20\%} & \multicolumn{1}{c}{\cellcolor{mygray}48\%} & \multicolumn{1}{c}{\cellcolor{mygray}64\%} \\
w/o Recovery & 2\% & 25\% & 23\% & 59\% & \multicolumn{1}{c}{\cellcolor{mygray}13\%} & \multicolumn{1}{c|}{\cellcolor{mygray}42\%} & 0\% & 16\% & 0\% & 16\% & 0\% & 21\% & \multicolumn{1}{c}{\cellcolor{mygray}0\%} & \multicolumn{1}{c|}{\cellcolor{mygray}18\%} & \multicolumn{1}{c}{\cellcolor{mygray}8\%} & \multicolumn{1}{c}{\cellcolor{mygray}33\%} \\
\midrule
% \multicolumn{17}{l}{\textit{Full Method}} \\
% \addlinespace[2pt]
\textbf{\nm (ours)} & 78\% & 88\% & \textbf{85\%} & \textbf{89\%} & \multicolumn{1}{c}{\cellcolor{mygray}\textbf{82\%}} & \multicolumn{1}{c|}{\cellcolor{mygray}\textbf{89\%}} & \textbf{53\%} & \textbf{79\%} & \textbf{37\%} & \textbf{84\%} & \textbf{43\%} & \textbf{74\%} & \multicolumn{1}{c}{\cellcolor{mygray}\textbf{44\%}} & \multicolumn{1}{c|}{\cellcolor{mygray}\textbf{79\%}} & \multicolumn{1}{c}{\cellcolor{mygray}\textbf{69\%}} & \multicolumn{1}{c}{\cellcolor{mygray}\textbf{86\%}} \\
\bottomrule
\end{tabular}%
}
\end{table*}

%% file: tables/perturbation.tex
\begin{figure}[t]
    \centering
    \begin{tikzpicture}
        \begin{axis}[
            width=0.9\linewidth,
            height=5.5cm,
            enlarge x limits=0.5,
            ymin=0, ymax=1.2,
            ytick={0, 0.2, 0.4, 0.6, 0.8, 1.0},
            ylabel={\textbf{Avg. Success Rate}},
            xtick={1, 2},
            xticklabels={Canonical, Noisy Init.},
            xticklabel style={font=\small, align=center},
            ylabel style={font=\small},
            ymajorgrids=true,
            grid style=dashed,
            legend style={
                at={(0.5, 0.95)},
                anchor=north,
                legend columns=2,
                font=\small,
                draw=none,
                fill=white, fill opacity=0.8,
                /tikz/every even column/.append style={column sep=0.4cm}
            },
            /pgf/number format/.cd, use period, 1000 sep={} 
        ]

            % --- BARS: LiLo-VLA ---
            \addplot[
                ybar, bar width=20pt, bar shift=-12pt,
                fill=orange, draw=none, opacity=0.4,
            ] coordinates {(1, 0.800) (2, 0.733)};

            % --- BARS: w/o Perturb ---
            \addplot[
                ybar, bar width=20pt, bar shift=12pt,
                fill=gray, draw=none, opacity=0.3,
            ] coordinates {(1, 0.741) (2, 0.578)};

            % --- LINES + ERROR BARS: LiLo-VLA ---
            \addplot[
                sharp plot, 
                xshift=-12pt, 
                color=orange!90!black, mark=*, line width=1.5pt,
                error bars/.cd, y dir=both, y explicit,
                error bar style={line width=1pt, solid}
            ]
            coordinates {
                (1, 0.800) += (0, 0.15) -= (0, 0.15)
                (2, 0.733) += (0, 0.14) -= (0, 0.14)
            };
            % 修改了这里：给文字加了颜色
            \addlegendentry{\textcolor{orange!90!black}{\textbf{LiLo-VLA} (Ours)}}

            % --- LINES + ERROR BARS: w/o Perturb ---
            \addplot[
                sharp plot, 
                xshift=12pt, 
                color=gray!80!black, mark=square*, line width=1.5pt, dashed,
                error bars/.cd, y dir=both, y explicit,
                error bar style={line width=1pt, solid}
            ]
            coordinates {
                (1, 0.741) += (0, 0.19) -= (0, 0.19)
                (2, 0.578) += (0, 0.18) -= (0, 0.18)
            };
            % 修改了这里：给文字加了颜色
            \addlegendentry{\textcolor{gray!80!black}{w/o Perturb}}

        \end{axis}
    \end{tikzpicture}
    \caption{\textbf{Impact of State Perturbation.} Average success rates across 27 unique skills demonstrate that our interaction policy remains robust to initial pose noise due to state perturbation, whereas the unperturbed policy degrades significantly.}
    \label{fig:ablation_perturbation}
\end{figure}

%% file: sections/real_experiments.tex
\section{Real Robot Experiments}
\label{sec:real_world}

\begin{figure*}[t]
    \centering
    \includegraphics[width=\linewidth]{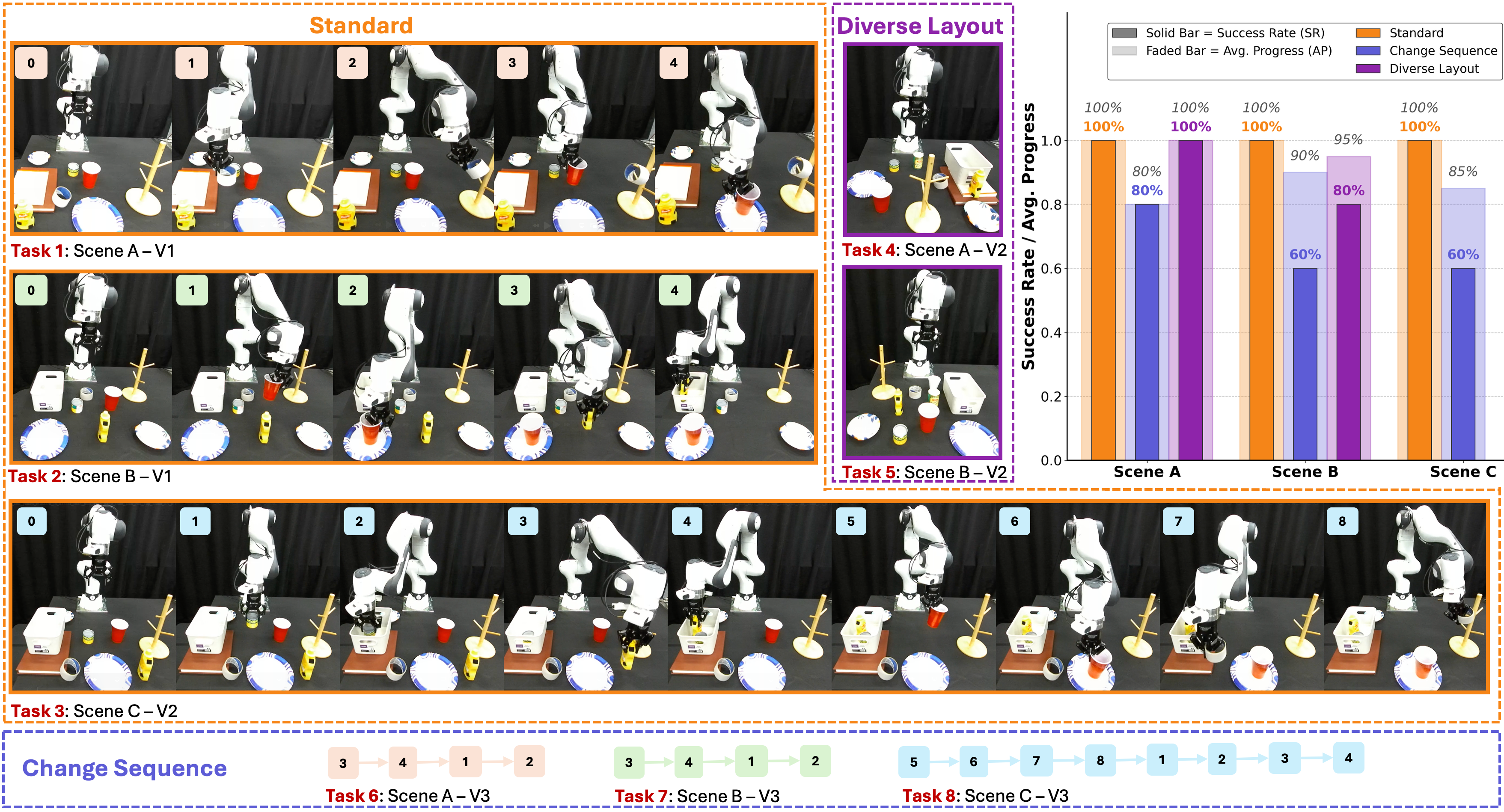} % Make sure this matches your file path
    \vspace{-15pt}
    \caption{\textbf{Real-World Experimental Evaluation.} We evaluate LiLo-VLA on 8 tasks ranging from 4 to 8 steps. To test generalization, we introduce: ``Standard'' configurations (Task 1-3), ``Diverse Layouts'' with more complex background and large layout variation (Task 4-5), and ``Change Sequence'' with permuted skill orders. The quantitative results (top right) demonstrate that LiLo-VLA maintains robust performance across all variations.}
    \label{fig:real_robot}
\end{figure*}

% \subsection{System Implementation and Perception Stack}
% \label{subsec:real_setup}
% % GOAL: Describe the hardware and software stack.
% % STRATEGY: Emphasize "Off-the-shelf" hardware (standard) and "Model-Agnostic" software (flexible).

% % 1. Hardware Setup
% % - Robot: Franka Emika Panda arm + Robotiq 2F-85 gripper.
% % - Cameras: Dual-camera setup following the DROID protocol [CITE].
% %   - 3rd Person: ZED 2 Stereo Camera (Global context/Reaching).
% %   - Wrist: ZED Mini (Object-centric manipulation).

% % 2. Perception Stack (From Oracle to Real)
% % - Object Detection/Segmentation: YOLOE [CITE] (specifically PP-YOLOE or YOLO-World depending on your implementation).
% % - Pose Estimation: FoundationPose [CITE] for 6D object pose to inform the Reaching Module.
% % - Data Collection: We collected a small set of teleoperated demonstrations for atomic skills only (no long-horizon demos).

% % 3. Strategic Backbone Switch (The "Flexibility" Argument)
% % - Mention that we switched the Interaction Policy backbone from OpenVLA-OFT (used in Sim) to Pi0.5 [CITE] in Real World.
% % - REASONING: This validates the model-agnostic nature of COVAL. It proves our framework enhances *any* capable VLA, not just one specific architecture.

\subsection{Real-World System Implementation}
\label{subsec:real_setup}

We implement \nm on a Franka Emika Panda robot equipped with a Robotiq 2F-85 parallel gripper. Following the standardized hardware protocol of DROID~\cite{khazatsky2024droid} we utilize a dual-camera setup where a static third-person ZED 2 stereo camera provides global context for the Reaching Module and a wrist-mounted ZED Mini camera captures egocentric data for the Interaction Module.

To transition from the oracle perception used in simulation to the unstructured real world we deploy a modular perception stack. We utilize YOLOE~\cite{wang2025yoloerealtimeseeing} for open-vocabulary object detection and segmentation coupled with FoundationPose~\cite{wen2024foundationpose} to estimate the 6D pose of target objects. This geometric information is ingested by the Reaching Module to generate collision-free motion plans to the target vicinity.

Complementing our simulation experiments which utilized OpenVLA-OFT~\cite{kim2025fine}, we deploy \nm with the Pi0.5~\cite{intelligence2504pi0} backbone in the real world. This architectural switch empirically validates the model-agnostic nature of our framework demonstrating that the modular benefits of \nm can enhance diverse VLA architectures without modification. To train this policy we collect a minimal set of teleoperated demonstrations covering only atomic skills without any long-horizon sequence data. Finally due to the severe occlusions and visual clutter inherent in our evaluation tasks, we employ a human-in-the-loop protocol for success verification of each skill.

\subsection{Evaluation Tasks: Stress-Testing Generalization}
\label{subsec:real_tasks}
% GOAL: Define the tasks and the rigorous testing protocols (Variants).
% REFERENCE: Refer to Fig. \ref{fig:real_robot} explicitly.

% 1. Task Definitions (The Core Tasks)
% - Task 1 (4 Skills): e.g., "Table Setting" (Pick cup, Place cup, etc.).
% - Task 2 (4 Skills): e.g., "Object Sorting".
% - Task 3 (8 Skills): "Ultra-Long Sequence" (Kitchen Organization).

% 2. Variant Protocols (The "Stress Test")
% - Standard: Canonical setup.
% - Variant 1 (Spatial/Visual Shift): Applied to Task 1 & 2.
%   - Drastic layout changes + Unseen visual clutter (distractors).
%   - Tests Object-Centricity and Reaching robustness.
% - Variant 2 (Temporal Shift): Applied to Task 1, 2, & 3.
%   - Permuted execution order (Change Sequence).
%   - Tests Compositionality (understanding skills vs. memorizing trajectories).

To evaluate the zero-shot generalization capabilities of \nm in the physical world, we design three long-horizon tasks of increasing complexity as illustrated in Fig. \ref{fig:real_robot}. Task 1 and Task 2 each consist of 4 atomic skills, while Task 3 represents an ultra-long sequence chaining 8 atomic skills to push the limits of temporal scalability.

Beyond the Standard evaluation where tasks are executed in canonical configurations, we introduce two variant protocols to stress-test specific generalization axes. The first protocol, \textit{Diverse Layout} (Variant 1), is applied to Tasks 1 and 2. Here we drastically alter the initial workspace layout and introduce previously unseen visual distractors. This setting evaluates the robustness of the Object-Centric VLA against visual clutter. The second protocol, \textit{Change Sequence} (Variant 2), is applied across all three tasks. We permute the execution order of the atomic skills forcing the system to generate novel long-horizon behaviors that were never seen during data collection. This protocol strictly tests compositional generalization distinguishing true skill understanding from trajectory memorization.

\subsection{Results Analysis}
\label{subsec:real_results}
% GOAL: Quantitative analysis using the Bar Chart in Fig. 6.
% KEY CLAIM: Complexity does not kill performance.

% 1. Quantitative Success
% - High absolute performance on Standard tasks (e.g., 100% SR for Task 1 & 3).
% - Shows the sim-to-real transfer is successful.

% 2. Robustness Verification (The "No Drop" Argument)
% - Compare Standard vs. Variant 1 (Diverse Layout):
%   - Result: Performance remains statistically comparable ($p > 0.05$).
%   - Insight: Proves Perception + Reaching handles spatial shifts, and Masking handles visual clutter.
% - Compare Standard vs. Variant 2 (Change Sequence):
%   - Result: Stable performance.
%   - Insight: Proves the system is compositional.

% 3. Conclusion
% - The combination of the new Backbone (Pi0.5) and our modular design works robustly in the real world.

We report the quantitative results in Fig. \ref{fig:real_robot} based on 5 evaluation trials per configuration. Under Standard conditions \nm achieves a 100\% success rate across all three tasks demonstrating effective sim-to-real transfer. When introduced to the ``Diverse Layout'' protocol the system exhibits strong resilience to spatial shifts and visual distractors maintaining a 100\% success rate on Task 1 and 80\% on Task 2. This stability validates the robustness of our object-centric Interaction Module against environmental noise. Furthermore under the ``Change Sequence'' protocol performance remains robust even when atomic skills are permuted into different orders. In the 8-step ultra-long Task 3 the system maintains a 60\% success rate with 85\% average progress. Collectively these results confirm that \nm enables robust real-world manipulation capable of handling severe visual clutter and flexible skill composition.

%% file: sections/conclusion.tex
\section{Conclusion and Limitations}
\label{sec:conclusion}

In this work, we introduced \nm, a modular framework that integrates a Reaching Module for global transport with an Interaction Module for local manipulation. By incorporating a closed-loop recovery mechanism our system effectively addresses observation space shifts and achieves robust performance in both simulation benchmarks and real-world deployments.

Despite these advancements, our reliance on external perception models introduces limitations where accurate detection of transparent or severely occluded objects remains challenging. Additionally, the overall system performance is bounded by the atomic proficiency of the underlying VLA backbone. Future work will investigate active perception strategies that autonomously navigate to the most favorable viewpoints for execution, thereby mitigating visual occlusions and maximizing the performance of the VLA policy.

%% file: suppl_sections/overview.tex
\section*{Overview}
This supplementary material provides comprehensive details supporting the main paper. 
Section~\ref{sec:add_exp_results} presents qualitative visualizations of our real-world experiments, highlighting specific mechanisms for robustness and failure recovery. 
Section~\ref{sec:bm_details} defines the full atomic skill library and details the exact task sequences for the LIBERO-Long++ and Ultra-Long benchmarks. 
Section~\ref{sec:ext_details_impl} elaborates on the relative pose representation designed to ensure zero-shot generalization to novel workspace configurations.
Section~\ref{sec:data_eff_cob} offers a theoretical analysis of the combinatorial complexity involved in long-horizon tasks, demonstrating the data efficiency of our modular approach. 
Section~\ref{train_impl_details} outlines the training objectives, hyperparameters, and the success checker heuristics utilized in our implementation. 
Finally, Section~\ref{sec:hardware_sys} describes the hardware specifications and perception models utilized in our real-world system.

%% file: suppl_sections/additional_experimental_results.tex
\section{Additional Experimental Results}
\label{sec:add_exp_results}
% GOAL: Visual proof of robustness and honest failure analysis.

\begin{figure*}[t]
    \centering
    % Replace 'images/real_world_experiments_full.png' with your actual filename
    \includegraphics[width=\linewidth]{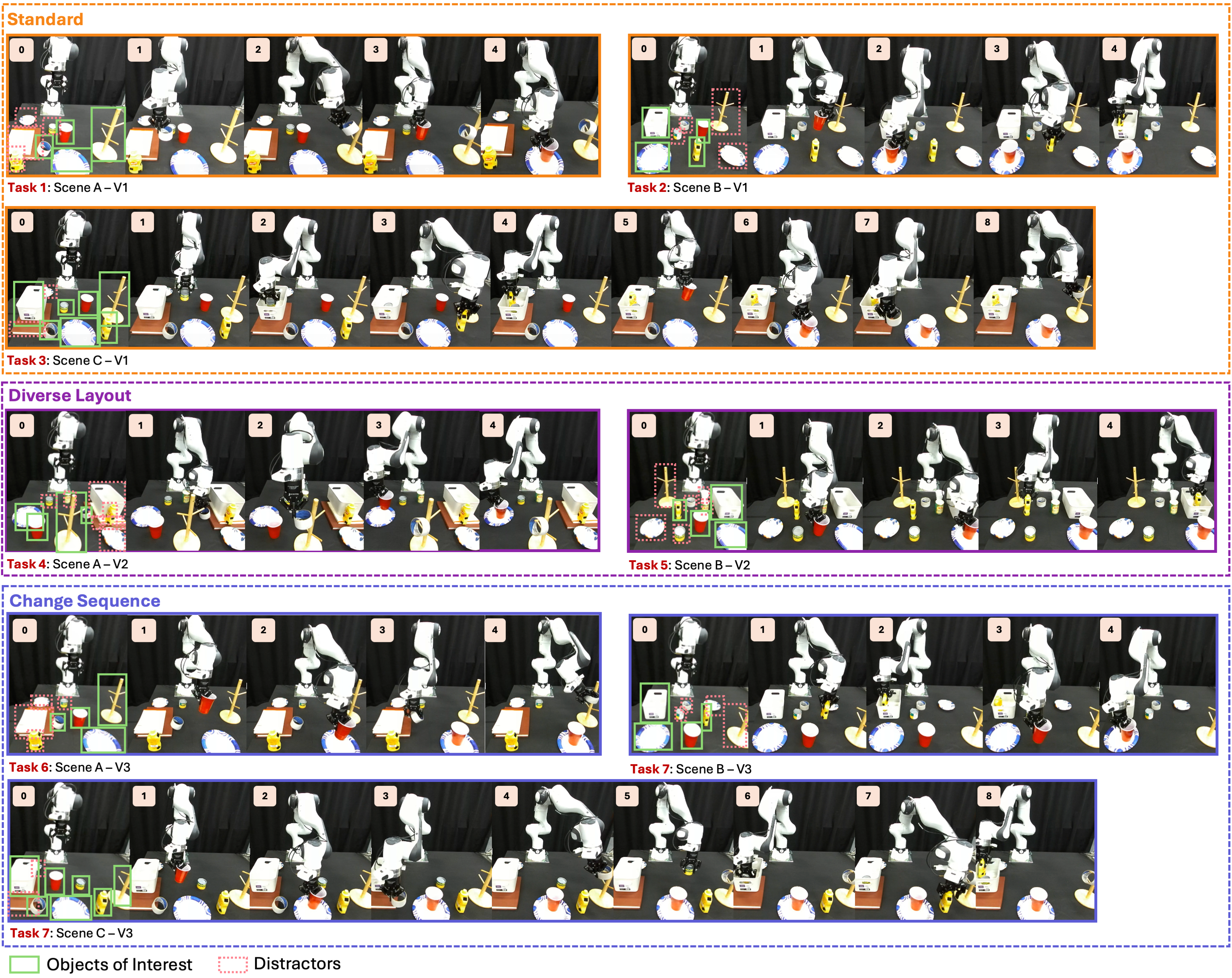}
    
    \caption{\textbf{Qualitative Real-World Experimental Results.}}
    \label{fig:real_world_results}
\end{figure*}

\subsection{Real-World Task Specifications}
To evaluate the compositional generalization of LiLo-VLA, we define a library of 8 unique atomic skills across three distinct scenes. Table~\ref{tab:atomic_skills} defines these primitives, while Table~\ref{tab:task_sequences} details the exact execution sequences for all 8 evaluation tasks.

\begin{table}[h]
\centering
\caption{Real-World Atomic Skill Library}
\label{tab:atomic_skills}
\small
\begin{tabular}{c l}
\hline
\textbf{Label} & \textbf{Object-Centric Instruction} \\ \hline
$S_1$ & ``pick the tape'' \\
$S_2$ & ``hang the tape'' \\
$S_3$ & ``pick the red cup'' \\
$S_4$ & ``place the red cup on the plate'' \\
$S_5$ & ``pick the yellow mustard'' \\
$S_6$ & ``place the yellow mustard in the basket'' \\
$S_7$ & ``pick the corn'' \\
$S_8$ & ``place the corn in the basket'' \\ \hline
\end{tabular}
\end{table}

\begin{table}[h]
\centering
\caption{Detailed Task Specifications for Real-World}
\label{tab:task_sequences}
\small
\begin{tabularx}{\linewidth}{l c l X}
\hline
\textbf{Task ID} & \textbf{Type} & \textbf{Steps} & \textbf{Skill Sequence} \\ \hline
Task 1 & Standard & 4 & $S_1 \to S_2 \to S_3 \to S_4$ \\
Task 2 & Standard & 4 & $S_3 \to S_4 \to S_5 \to S_6$ \\
Task 3 & Standard & 8 & $S_7 \to S_8 \to S_5 \to S_6 \to S_3 \to S_4 \to S_1 \to S_2$ \\ \hline
Task 4 & Clutter & 4 & $S_1 \to S_2 \to S_3 \to S_4$ \\
Task 5 & Clutter & 4 & $S_3 \to S_4 \to S_5 \to S_6$ \\ \hline
Task 6 & Permuted & 4 & $S_3 \to S_4 \to S_1 \to S_2$ \\
Task 7 & Permuted & 4 & $S_5 \to S_6 \to S_3 \to S_4$ \\
Task 8 & Permuted & 8 & $S_3 \to S_4 \to S_1 \to S_2 \to S_7 \to S_8 \to S_5 \to S_6$ \\ \hline
\end{tabularx}
\end{table}

\subsection{Real-World Rollout Visualizations}
We provide comprehensive visualizations of our real-world experiments in Fig.~\ref{fig:real_world_results}. The top panel displays successful execution rollouts for all 8 evaluation tasks, where we explicitly mark the active object of interest in green and all distractors in red. Note that for any specific atomic skill, there is only one target object; consequently, all other objects in the scene, including targets for other skills, are treated as distractors, highlighting the robustness of our object-centric policy against clutter.

\begin{figure*}[t]
    \centering
    \includegraphics[width=\linewidth]{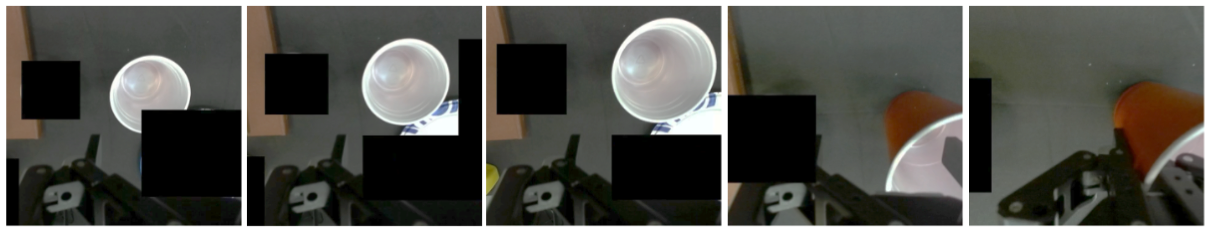}
    \caption{\textbf{Visual Robustness.} Random masking on wrist camera inputs forces the policy to focus strictly on the target object effectively ignoring background clutter.}
    \label{fig:wrist}
\end{figure*}

\begin{figure*}[t]
    \centering
    \includegraphics[width=\linewidth]{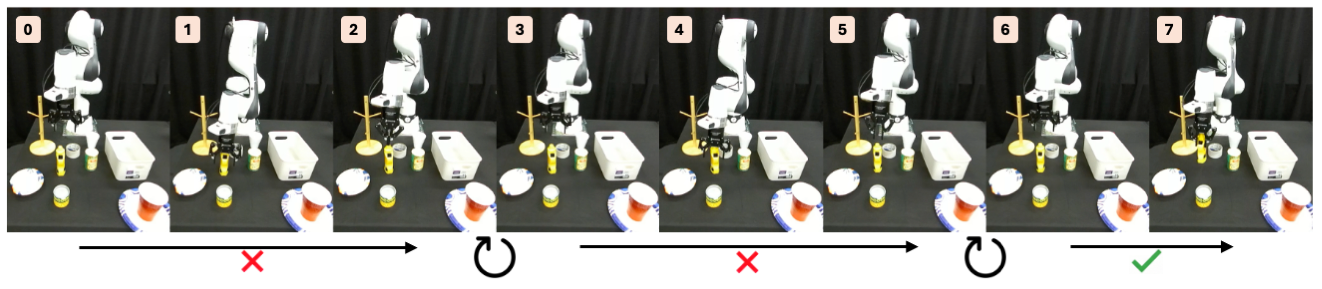}    
    \caption{\textbf{Recovery Mechanism.} Upon detecting a grasp failure, the system autonomously re-estimates the object pose and resets the arm to retry the execution.}
    \label{fig:recovery}
\end{figure*}

To explain LiLo-VLA's robustness, we visualize two core mechanisms. First, for perception, Fig.~\ref{fig:wrist} shows the wrist camera inputs. We apply random black masks to these images, which forces the policy to ignore background clutter and focus only on the target object. Second, for execution, Fig.~\ref{fig:recovery} shows the recovery process. After grasp failures in Frames 1 and 4, the system re-estimates the object's new pose and uses the Reaching Module to reset the arm to the approach pose (Frames 3 and 6). This allows the Interaction Module to retry the skill, eventually leading to a success (Frame 7).

%% file: suppl_sections/benchmark_details.tex
\section{Benchmark Details}
\label{sec:bm_details}

\subsection{Atomic Skill Library}
We define a comprehensive library of 22 atomic skills used across LIBERO-Long++ and Ultra-Long suites. To facilitate concise task specification, we assign a unique label to each skill in Table~\ref{tab:sim_skill_lib}.

\begin{table}[h]
\centering
\caption{Simulation Atomic Skill Library (LIBERO-Long++ and Ultra-Long)}
\label{tab:sim_skill_lib}
\scriptsize
\begin{tabularx}{\linewidth}{l X | l X}
\hline
\textbf{ID} & \textbf{Skill Description} & \textbf{ID} & \textbf{Skill Description} \\ \hline
$S_1$ & Pick Alphabet Soup & $S_{12}$ & Pick Moka Pot \\
$S_2$ & Place Alphabet Soup in Basket & $S_{13}$ & Place Moka Pot on Stove \\
$S_3$ & Pick Cream Cheese & $S_{14}$ & Turn On Stove \\
$S_4$ & Place Cream Cheese in Basket & $S_{15}$ & Pick Butter \\
$S_5$ & Pick Tomato Sauce & $S_{16}$ & Place Butter in Basket \\
$S_6$ & Place Tomato Sauce in Basket & $S_{17}$ & Pick Chocolate Pudding \\
$S_7$ & Pick Black Bowl & $S_{18}$ & Place Chocolate Pudding Right of Plate \\
$S_8$ & Place Black Bowl on Plate & $S_{19}$ & Pick White Mug \\
$S_9$ & Stack Black Bowl on Black Bowl & $S_{20}$ & Place White Mug on Plate \\
$S_{10}$ & Place Black Bowl in Bottom Drawer & $S_{21}$ & Pick Yellow and White Mug \\
$S_{11}$ & Close Bottom Drawer & $S_{22}$ & Place Yellow and White Mug on Right Plate \\
\hline
\end{tabularx}
\end{table}

\subsection{Detailed Task Specifications}
We provide the exact atomic skill sequences for all evaluation tasks in Table~\ref{tab:sim_task_specs}. All skill IDs (e.g., $S_1, S_{12}$) correspond to the definitions in the Atomic Skill Library (Table~\ref{tab:sim_skill_lib}).

For Suite 1 (LIBERO-Long++), we define 6 core tasks focused on robustness and reordering. Each task consists of a Standard sequence and a corresponding Variant, where the latter permutes the execution order of atomic skills to evaluate the policy's zero-shot compositional generalization.

For Suite 2 (Ultra-Long), we define 3 complex long-horizon scenarios to evaluate extreme temporal scalability. Each scenario includes one Standard sequence and two additional Variant sequences. Notably, the ``Table Organization'' task extends up to 16 steps, serving as a rigorous stress test for the system's ability to maintain coherent execution over extended horizons.

\begin{table*}[t]
\centering
\caption{Detailed Task Specifications for Simulation Benchmarks}
\label{tab:sim_task_specs}
\scriptsize
\begin{tabularx}{\textwidth}{l l c X}
\hline
\textbf{Suite} & \textbf{Task Name} & \textbf{Steps} & \textbf{Skill Sequence (IDs from Table~\ref{tab:sim_skill_lib})} \\ \hline
\rowcolor[gray]{0.95} \multicolumn{4}{l}{\textbf{Suite 1: LIBERO-Long++ (Robustness and Reordering)}} \\
Standard & T1: Stove and Moka & 3 & $S_{14} \to S_{12} \to S_{13}$ \\
Variant & T2: Stove and Moka & 3 & $S_{12} \to S_{13} \to S_{14}$ \\ \hline
Standard & T3: Soup and Cheese & 4 & $S_{1} \to S_{2} \to S_{3} \to S_{4}$ \\
Variant & T4: Soup and Cheese & 4 & $S_{3} \to S_{4} \to S_{1} \to S_{2}$ \\ \hline
Standard & T5: Soup and Tomato & 4 & $S_{1} \to S_{2} \to S_{5} \to S_{6}$ \\
Variant & T6: Soup and Tomato & 4 & $S_{5} \to S_{6} \to S_{1} \to S_{2}$ \\ \hline
Standard & T7: Cheese and Butter & 4 & $S_{3} \to S_{4} \to S_{15} \to S_{16}$ \\
Variant & T8: Cheese and Butter & 4 & $S_{15} \to S_{16} \to S_{3} \to S_{4}$ \\ \hline
Standard & T9: Two Mugs & 4 & $S_{19} \to S_{20} \to S_{21} \to S_{22}$ \\
Variant & T10: Two Mugs & 4 & $S_{21} \to S_{22} \to S_{19} \to S_{20}$ \\ \hline
Standard & T11: Mug and Pudding & 4 & $S_{19} \to S_{20} \to S_{17} \to S_{18}$ \\
Variant & T12: Mug and Pudding & 4 & $S_{17} \to S_{18} \to S_{19} \to S_{20}$ \\ \hline

\rowcolor[gray]{0.95} \multicolumn{4}{l}{\textbf{Suite 2: Ultra-Long (Scalability and Compositionality)}} \\
Standard & T13: Kitchen Organization & 9 & $S_{7}\to S_{8}\to S_{7}\to S_{8}\to S_{7}\to S_{9}\to S_{3}\to S_{4}\to S_{11}$ \\
Variant 1 & T14: Kitchen Org. (V2) & 9 & $S_{3}\to S_{4}\to S_{7}\to S_{8}\to S_{7}\to S_{9}\to S_{7}\to S_{8}\to S_{11}$ \\
Variant 2 & T15: Kitchen Org. (V3) & 9 & $S_{7}\to S_{8}\to S_{7}\to S_{9}\to S_{3}\to S_{4}\to S_{7}\to S_{8}\to S_{11}$ \\ \hline

Standard & T16: Cooking Preparation & 10 & $S_{14}\to S_{12}\to S_{13}\to S_{14}\to S_{12}\to S_{13}\to S_{7}\to S_{8}\to S_{7}\to S_{8}$ \\
Variant 1 & T17: Cooking Prep. (V2) & 10 & $S_{7}\to S_{8}\to S_{7}\to S_{8}\to S_{14}\to S_{12}\to S_{13}\to S_{14}\to S_{12}\to S_{13}$ \\
Variant 2 & T18: Cooking Prep. (V3) & 10 & $S_{7}\to S_{8}\to S_{14}\to S_{12}\to S_{13}\to S_{7}\to S_{8}\to S_{14}\to S_{12}\to S_{13}$ \\ \hline

Standard & T19: Table Organization & 16 & $S_{5}\to S_{6}\to S_{5}\to S_{6}\to S_{1}\to S_{2}\to S_{1}\to S_{2}\to S_{3}\to S_{4}\to S_{3}\to S_{4}\to S_{7}\to S_{10}\to S_{7}\to S_{8}$ \\
Variant 1 & T20: Table Org. (V2) & 16 & $S_{1}\to S_{2}\to S_{5}\to S_{6}\to S_{1}\to S_{2}\to S_{5}\to S_{6}\to S_{3}\to S_{4}\to S_{3}\to S_{4}\to S_{7}\to S_{10}\to S_{7}\to S_{8}$ \\
Variant 2 & T21: Table Org. (V3) & 16 & $S_{1}\to S_{2}\to S_{1}\to S_{2}\to S_{5}\to S_{6}\to S_{5}\to S_{6}\to S_{3}\to S_{4}\to S_{3}\to S_{4}\to S_{7}\to S_{10}\to S_{7}\to S_{8}$ \\ \hline
\end{tabularx}
\end{table*}

%% file: suppl_sections/system_design_details.tex
\section{Extended Methodology Details}
\label{sec:ext_details_impl}

While the main text details our object-centric visual processing, the handling of proprioception is equally critical for generalization. To further enforce an object-centric inductive bias in our simulation backbone (OpenVLA-OFT), we transform the standard absolute end-effector pose into a relative frame. Specifically, the policy input is computed as $T_{obj}^{ee} = (T_{world}^{obj})^{-1} T_{world}^{ee}$, where $T_{world}^{ee}$ and $T_{world}^{obj}$ represent the global poses of the end-effector and target object, respectively. This formulation renders the interaction policy invariant to global workspace shifts. In contrast, our real-world Pi0.5 policy operates directly in joint space and thus does not utilize this Cartesian transformation. We empirically validate the robustness of this design by introducing random spatial perturbations of up to 20cm (where the object's position is randomly placed within this range); the system maintains strong performance with no statistically significant degradation (overall success rate over all 22 atomic skills in \ref{tab:sim_skill_lib} dropped only slightly from 0.80 to 0.75, $p > 0.05$).

%% file: suppl_sections/data_efficiency_and_scalability_analysis.tex
\section{Data Efficiency and Combinatorial Complexity}
\label{sec:data_eff_cob}

To rigorously feature the data efficiency advantage of LiLo-VLA over end-to-end baselines, we analyze the sample complexity required to generalize to novel long-horizon task sequences. Let a long-horizon task $\mathcal{T}$ be composed of a set of atomic skills $\mathcal{S} = \{s_1, s_2, \dots, s_N\}$. In a Task and Motion Planning (TAMP) formulation, the valid execution orders are governed by causal dependencies, where the effect of a skill $s_i$ serves as a precondition for a subsequent skill $s_j$. These dependencies induce a partial ordering over $\mathcal{S}$, denoted as $\prec$.

\textbf{End-to-End Complexity.} An end-to-end policy $\pi_{e2e}$ must implicitly internalize these valid orderings from demonstration data. To achieve robust compositional generalization, the policy must observe sufficient coverage of the set of all valid linearizations (i.e., topological sorts) consistent with the partial order $\prec$. For a task involving $M$ independent objects, each requiring a pick-place sequence, the number of valid linearizations can be as large as $\Theta\left(\frac{(2M)!}{2^M}\right)$ when objects have no inter-dependencies. For our ultra-long tasks ($N = 16$, $M = 8$), this combinatorial explosion renders it intractable to cover the distribution of valid trajectories via demonstrations alone.

\textbf{Modular Complexity (LiLo-VLA).} In contrast, LiLo-VLA decouples the task into independent atomic execution units. Since our Interaction Module $\pi_{int}$ is strictly object-centric and conditioned only on the immediate target, the learning problem reduces to mastering the set of unique atomic skills $\mathcal{U} \subseteq \mathcal{S}$. Consequently, the data requirement scales linearly, $\mathcal{O}(|\mathcal{U}|)$, independent of the total horizon length $N$ or the combinatorial complexity of the task structure. For the 16-step ``Table Organization'' task, this reduces the learning problem from covering millions of potential trajectory variations to simply mastering the 9 unique atomic primitives listed in Table~\ref{tab:atomic_skills}.

%% file: suppl_sections/training_and_implementation_details.tex
\section{Training and Implementation Details}
\label{train_impl_details}

\subsection{Training Objectives}
We employ two distinct training objectives corresponding to the different backbones used in our simulation and real-world experiments.

\textbf{OpenVLA-OFT (Simulation).} For our simulation benchmarks, we utilize OpenVLA-OFT~\cite{kim2025fine}, which departs from the standard discrete tokenization used in original VLA models. Instead, it adopts a \textit{continuous action representation} and employs an Optimized Fine-Tuning (OFT) recipe that integrates parallel decoding and action chunking. Consequently, the model is trained via a regression objective rather than classification. We minimize the L1 loss between the predicted action chunk $\hat{\mathbf{a}}_{t:t+H}$ and the ground-truth action sequence $\mathbf{a}_{t:t+H}$:
\begin{equation}
    \mathcal{L}_{OFT} = \frac{1}{H} \sum_{k=0}^{H-1} \left\| \hat{\mathbf{a}}_{t+k} - \mathbf{a}_{t+k} \right\|_1
\end{equation}
where $H$ is the action chunk size, and the model predicts the entire chunk in parallel conditioned on the observation $o_t$ and instruction $l$.

\textbf{Pi0.5 (Real-World).} For real-world validation, we adopt the Pi0.5 backbone~\cite{intelligence2504pi0}, which models the continuous action distribution using Conditional Flow Matching (CFM). The model learns a vector field $v_\theta$ that transports a Gaussian noise distribution $x_0 \sim \mathcal{N}(0, I)$ to the data distribution $x_1$ (ground-truth actions) over a virtual time $\tau \in [0, 1]$. The training objective minimizes the mean squared error between the predicted vector field and the target flow:
\begin{equation}
    \mathcal{L}_{FM} = \mathbb{E}_{\tau, x_0, x_1} \left[ || v_\theta(\phi_\tau(x_0, x_1), \tau, o_t, l) - (x_1 - x_0) ||^2 \right]
\end{equation}
where $\phi_\tau(x_0, x_1) = (1 - \tau)x_0 + \tau x_1$ represents the linear interpolation path.

\begin{figure*}[t]
    \centering
    \includegraphics[width=0.8\linewidth]{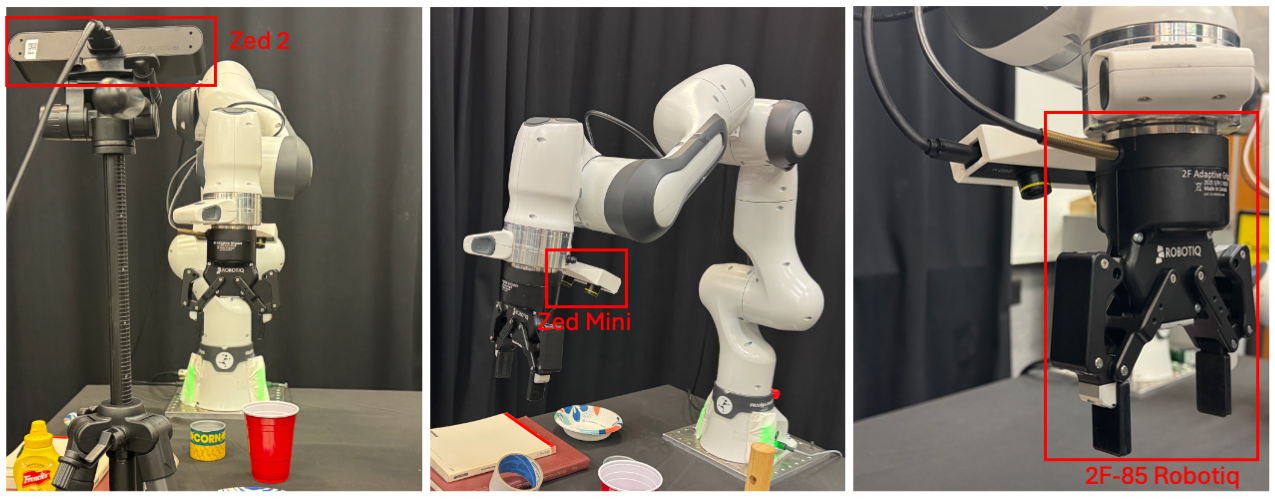} 
    \caption{\textbf{Real-world hardware setup}. Our system consists of a Franka Emika Panda arm, a wrist-mounted ZED mini camera for fine-grained interaction, and a fixed ZED 2i camera for global reaching. Compute is distributed between a local workstation (Perception/Control) and a remote cluster (VLA Inference).}
    \label{fig:hardware_setup}
\end{figure*}

\subsection{Training Hyperparameters}
We train a single multi-task policy for each domain: one OpenVLA-OFT model across the simulation atomic skill library and one Pi0.5 model across the real-world atomic skill library. Table~\ref{tab:hyperparameters} provides a comprehensive overview of the hyperparameters used for these trainings.

\begin{table}[h]
\centering
\caption{Hyperparameters and Training Configuration.}
\label{tab:hyperparameters}
\small
\begin{tabularx}{\linewidth}{l X X}
\hline
\textbf{Hyperparameter} & \textbf{Simulation (OpenVLA-OFT)} & \textbf{Real-World (Pi0.5)} \\ \hline
Base Model & OpenVLA-7B & Pi0.5 (DROID) \\
Action Representation & Continuous (OFT) & Continuous (Flow Matching) \\
Finetuning Method & LoRA & Full \\
Action Chunk Size ($H$) & 8 & 16 \\
Input Observation & Wrist RGB + Proprioception & Wrist RGB + Proprioception \\
Training Objective & L1 Regression & Flow Matching \\
Optimizer & AdamW & AdamW \\
Learning Rate & $5 \times 10^{-4}$ & $5 \times 10^{-5}$ \\
LR Schedule & Cosine Decay & Cosine Decay \\
Batch Size & 64 & 32 \\
Training Steps & 100,000 & 30,000 \\
Random Erasing & Enabled  & Enabled \\
\hline
\end{tabularx}
\end{table}

\subsection{Success Checker Heuristic Functions}
In our simulation experiments, the verification function $\mathcal{V}(a_i)$ leverages ground-truth state information to evaluate skill execution. We primarily adopt the standard success predicates provided by the LIBERO benchmark. We create one success condition for the Pick skill: the skill is deemed successful only if the target object's vertical position ($z$-axis) increases by at least 3 cm relative to its pre-execution state. For all other atomic skills, we retain the default LIBERO success conditions without modification.

%% file: suppl_sections/infra_assets.tex
\section{Hardware and System Setup}
\label{sec:hardware_sys}

\subsection{Hardware Specifications}
Our real-world experimental setup follows the DROID hardware configuration standards. We use a 7-DoF Franka Emika Panda robotic arm equipped with a standard parallel jaw gripper (Robotiq 2F-85). The robot is operated using a joint impedance controller at a control frequency of 15 Hz. The vision system comprises two stereo cameras: a wrist-mounted Stereolabs ZED Mini for local object-centric observations and a single fixed third-person Stereolabs ZED 2 camera for global reaching. The VLA policy inference runs on a remote cluster node equipped with NVIDIA RTX A6000 GPUs, while all perception modules and real-time control loops operate on a local workstation equipped with a single NVIDIA RTX 4080 GPU. We visualize the complete physical setup in Fig.~\ref{fig:hardware_setup}.

\subsection{Perception Models}
We utilize FoundationPose~\cite{wen2024foundationpose} for 6D pose estimation and YOLOE~\cite{wang2025yoloerealtimeseeing} for 2D object detection and segmentation. Both models operate in real-time to enable closed-loop feedback. FoundationPose tracks the target object pose using an RGB image, an instance segmentation mask, and a 3D CAD model. We obtain the necessary 3D meshes by scanning the physical objects with the AR Code mobile application. For object detection, we employ YOLOE with visual prompting. Instead of conditioning detection on natural language text descriptions, we provide a reference image with a bounding box of the target object. We find this visual prompting approach offers higher robustness against background clutter compared to open-vocabulary text prompts.